\documentclass[lettersize,journal]{IEEEtran}
\usepackage{amsmath,amsfonts}
\usepackage{algorithmic}
\usepackage{algorithm}
\usepackage{array}
\usepackage[caption=false,font=normalsize,labelfont=sf,textfont=sf]{subfig}
\usepackage{textcomp}
\usepackage{stfloats}
\usepackage{url}
\usepackage{verbatim}
\usepackage{graphicx}
\usepackage{cite}
\usepackage[table]{xcolor}
\usepackage{booktabs}
\usepackage{multirow}
\usepackage{pifont}
\newcommand{\cmark}{\ding{51}}
\newcommand{\xmark}{\ding{55}}
\usepackage[colorlinks=true, linkcolor=blue, citecolor=blue, urlcolor=blue]{hyperref}
\begin{document}

\title{MSP-Net: Manifold-Guided Spectral Prompt Network for Hyperspectral Object Tracking}

\author{Juliu Li, Hanlin Qin, Shuowen Yang, Jingjing Li, Yuedong Tan, Shuai Yuan, Huixin Zhou

\thanks{Manuscript received April 19, 2021; revised August 16, 2026.}}

\markboth{Journal of \LaTeX\ Class Files,~Vol.~14, No.~8, August~2021}%
{Shell \MakeLowercase{\textit{et al.}}: A Sample Article Using IEEEtran.cls for IEEE Journals}


\maketitle

\begin{abstract}
Hyperspectral object tracking leverages abundant spectral information to provide unique advantages for target discrimination in complex scenes. However, existing methods typically treat hyperspectral images as multi-channel extensions of RGB images, performing feature fusion in fixed band order. This approach leads to models dependent on specific sensor configurations while neglecting manifold relationships between bands, making generalization to heterogeneous sensors difficult. Moreover, the discriminative contribution of bands dynamically changes with target attributes and scene variations, further limiting the representational capacity of static fusion strategies. To address this, we propose the Manifold-Guided Spectral Prompt Network (MSP-Net). This network first reconstructs band relationships and forms adaptive spectral grouping through graph-driven manifold routing, then jointly integrates grouped spectral statistics with template appearance to construct target-related dynamic conditional prompts, enhancing target features while suppressing background interference. Furthermore, as tracking progresses, spectral conditions continuously evolve based on intermediate target representations, enabling target prompts to adapt in real-time to appearance and scene changes. Meanwhile, reliable historical states are used to constrain target localization and scale fluctuations, significantly improving temporal stability in cross-sensor tracking. Experiments on HOT2020 and HOT2023 demonstrate that MSP-Net achieves AUC and Precision exceeding 0.80 and 0.96, respectively, exhibiting exceptional robustness under heterogeneous sensors, target deformation, and complex background conditions. The code will be released at https://github.com/GGML668897/MSP-Net.
\end{abstract}


\section{Introduction}
\IEEEPARstart{V}{isual} object tracking plays an important role in applications such as video surveillance, medical diagnosis, autonomous driving, smart agriculture, and intelligent early-warning systems \cite{DRSST,2,HyperTrack}. 

\begin{figure}[htbp]
\centering
\includegraphics[width=\columnwidth]{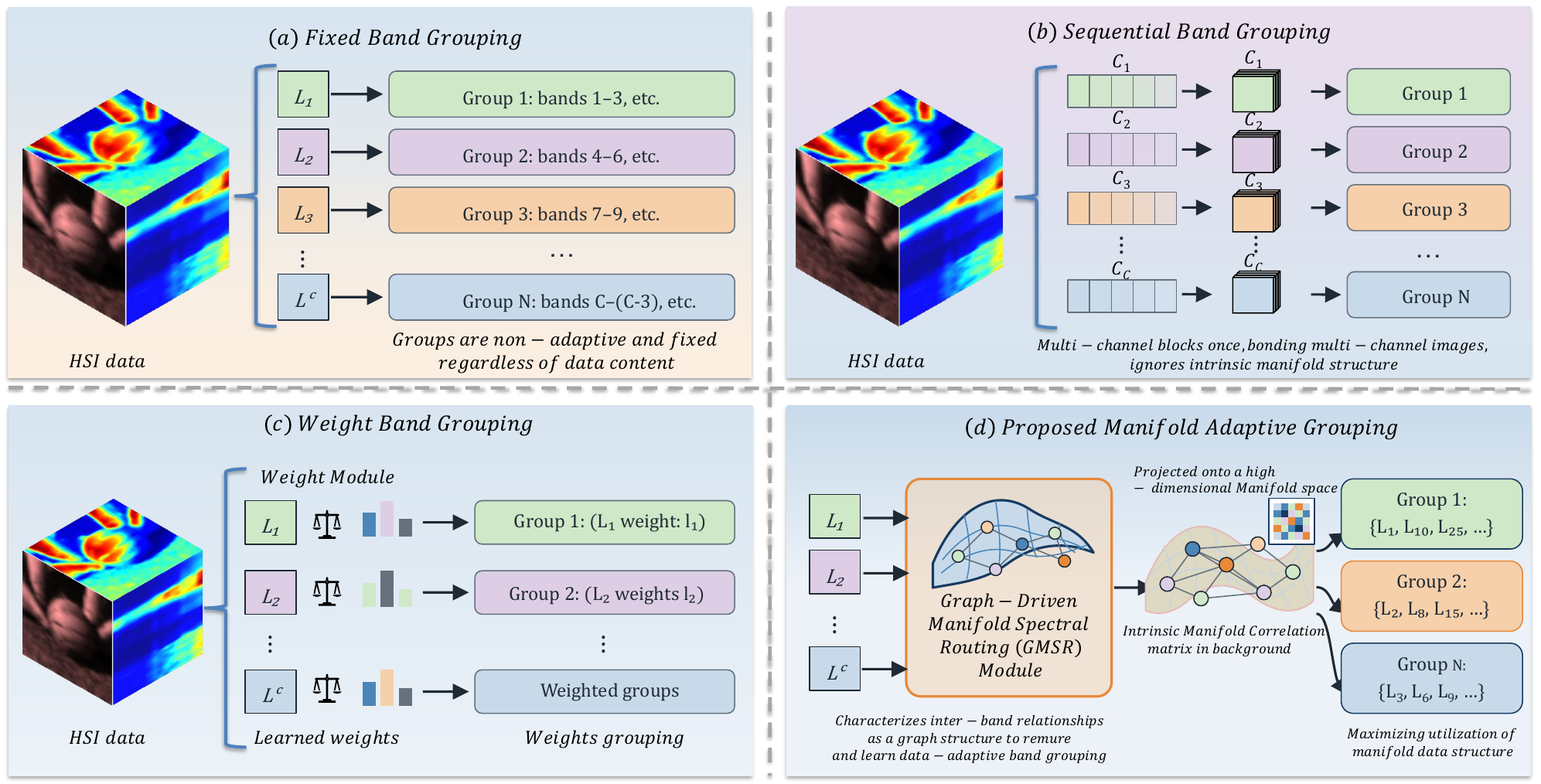}
\caption{Comparison between the existing (a) fixed grouping dimensionality reduction, (b) spectral segment sorting grouping dimensionality reduction, (c) weight grouping dimensionality reduction paradigms and the proposed (d) adaptive manifold paradigm.}
\label{fig:1}
\end{figure}

However, RGB-based trackers often struggle with strongly cluttered backgrounds\cite{SP-KAN}, similar distractors, deformation, and occlusion, leading to tracking drift\cite{4,CASS}. The early mainstream RGB tracking methods developed from correlation tracking methods, such as KCF\cite{KCF} and BACF\cite{BACF}, then advanced to Siamese networks \cite{SiamFC} incorporating deep learning, and further developed into transformer-based methods like OSTrack\cite{Ostrack}, MixFormer\cite{Mixformer}, and STARK\cite{STARK}, all of which have made leading research for target tracking. But relying solely on RGB information as a tracking driver cannot effectively alleviate tracking drift problems in scenarios such as target deformation and occlusion. 

Unlike RGB tracking, hyperspectral imaging provides wavelength-dependent material responses that support more reliable target identification in complex scenes\cite{MHT,MMF-Net,yang2025variational}. By exploiting spectral–spatial–temporal information, hyperspectral object tracking methods can be broadly divided into two categories. One is full-spectrum information perception, such as  CHP\cite{CHP} with full-band mapping, HHTrack\cite{HHTrack} with raw hyperspectral cubes, and E2E-MPT\cite{E2E-MPT} with embedded spectral unmixing for material-abundance estimation. The other is spectral dimensionality reduction, such as SEE-Net\cite{SEE-Net} and SiamBAG\cite{SiamBAG}, typically converting hyperspectral images into several three-channel pseudo-RGB groups for feature extraction and fusion. This strategy facilitates the use of pretrained RGB trackers and has been adopted by COALT\cite{COALT} through spectral band select, HyA-T\cite{HyA-T} through color-matching functions, and TBRNet\cite{TBRNet} through dynamic spectral projection and pseudo-color reorganization.

The two mainstream paradigms reveal a fundamental trade-off between spectral fidelity and computational efficiency: full-spectrum modeling preserves richer information but suffers from redundancy and high computational cost, whereas dimensionality reduction facilitates real-time inference and RGB-model transfer at the risk of losing discriminative spectral cues. Considering practical deployment, we adopt spectral dimensionality reduction but shift the focus from simple compression to preserving intrinsic inter-band structures. As illustrated in Fig.~\ref{fig:1}, conventional fixed, sequential, weighted, and similarity-based grouping strategies rely on rigid partitions that may disrupt spectral continuity and topology, motivating our manifold-adaptive grouping scheme. However, effective spectral grouping alone is insufficient for long-term tracking: evolving illumination, deformation, and occlusion can gradually invalidate spectral conditions, while static scale penalties struggle to balance deformation adaptation against drift suppression. These coupled challenges motivate the joint modeling of manifold-aware spectral representations, temporal condition evolution, and adaptive tracking stabilization.

Thus, we propose the Manifold-Guided Spectral Prompt Network (MSP-Net), which jointly models inter-band structures, target-conditioned information, and temporal states. First, the Graph-Driven Manifold Spectral Routing (GMSR) module constructs a band-relation graph and performs adaptive band routing and group-level aggregation according to content correlations. The routing strategy derived from the search region is consistently applied to both the template and search branches, ensuring aligned spectral representations. Based on the grouped features, the Decoupled Spectral-Condition Evolution (DSCE) strategy integrates group-level spectral statistics, template appearance information, and group-level weights to construct a joint spectral–target condition. During inference, this condition is periodically updated using intermediate template features. Subsequently, the Dual-Stream Spectral-Conditioned Prompt Modulation (DSCPM) mechanism generates target-aware prompts from the resulting condition. Through cross-attention and cosine-similarity gating, DSCPM selectively enhances the template and search features at successive Transformer layers, improving discrimination against similar background distractors. In addition, the Temporal Morphology Memory-Adaptive (TMMA) mechanism serves as an independent inference-stage post-processing module. It adaptively calibrates tracking predictions using historical states, spatial displacements, and morphological variations, thereby improving localization stability under substantial deformation and rapid motion. MSP-Net is systematically evaluated on four hyperspectral object-tracking benchmarks—HOT2020, HOT2023, HOT2024, and IMEC25. The experimental results demonstrate the effectiveness and generalization capability of the proposed method. The main contributions are summarized as follows:

(1) MSP-Net is proposed as a unified hyperspectral tracking framework that combines manifold-aware spectral modeling with dynamic prompt evolution to reduce spectral information loss, feature lag, and tracking drift.

(2) GMSR and DSCPM enhance spectral representation: GMSR performs graph-based routing in a nonlinear manifold space, while DSCPM exploits complementary spectral and appearance cues for stronger target-background discrimination.

(3) DSCE and TMMA improve temporal adaptation and localization stability. DSCE updates spectral conditions without parameter optimization, whereas TMMA adaptively regulates spatial and morphological variations. Experiments on hyperspectral datasets demonstrate strong robustness to deformation, illumination changes, and background interference.

\section{Relate Work}
This chapter discusses the related work of Hyperspectral tracking, including hyperspectral object tracking, dynamic routing and MoE, manifold learning and prompt learning.
\subsection{Hyperspectral Object Tracking}

Hyperspectral object tracking has evolved from correlation-filter methods to deep Siamese and Transformer-based frameworks. Early trackers such as DeepHKCF\cite{DeepHKCF}, CNHT\cite{CNHT}, and MHT\cite{MHT} incorporated deep features, spectral--spatial cues, and material representations into correlation filtering, but remained vulnerable to deformation, illumination changes, and background interference. Subsequent methods, including BAE-Net\cite{BAE-Net}, SEE-Net\cite{SEE-Net}, HA-Net\cite{HA-Net}, SiamBAG\cite{SiamBAG}, and SiamHYPER\cite{SiamHYPER}, improved spectral discrimination through band attention, ensemble learning, and Siamese architectures. More recently, HHTrack\cite{HHTrack}, HyA-T\cite{HyA-T}, SGAT\cite{SGAT}, and E2E-MPT\cite{E2E-MPT} have introduced hybrid attention, lightweight adapters, spectral gating, and material prompts to strengthen feature extraction and fusion. Despite this progression, many existing methods still rely on band selection, fixed grouping, pseudo-color transformation, or computationally intensive fusion, which may disrupt inter-band relationships and provide limited adaptation to the continuous spectral and appearance evolution of targets in long sequences.

Despite recent progress, multi-branch attention and complex fusion increase computational costs, while reliance on fixed spectral compression and limited template updates makes it difficult to follow continuous changes in target appearance, spectral response, and motion. To overcome these, our method adaptively routes bands within a nonlinear spectral-manifold space and dynamically evolves target representations over time, jointly preserving intrinsic spectral relationships and tracking temporal state variations in long sequences.

\subsection{Dynamic Routing and MoE}
As a core paradigm of conditional computation, the Mixture-of-Experts (MoE) architecture\cite{MoE} dynamically activates sparse subnetworks through a gating network, thereby substantially increasing model capacity while keeping computational cost relatively stable. This mechanism first achieved notable success in natural language processing. Representative models such as Switch Transformer\cite{SwitchTransformer},V-MoE\cite{V-MoE} and DeepSeekMoE\cite{DeepSeekMoE} significantly improved the training efficiency and generalization ability of large-scale models through simplified routing strategies and shared-expert designs. In recent years, MoE has also been extended to vision and multimodal learning. For example, MoCLE\cite{MoCLE} incorporates a gating mechanism into Transformer attention layers and activates experts according to instruction clusters, improving generalization to unseen data while maintaining computational efficiency.

In the challenging task of hyperspectral object tracking, the application of MoE architectures has gradually emerged as a promising research direction. HoT-MoE\cite{HOT-MOE} and FSMT\cite{FSMT} was the pioneer introduce MoE into hyperspectral tracking, demonstrating its potential for modeling high-dimensional structured features and supporting cross-modal adaptation. More recently, DRSST-Net\cite{DRSST}, built upon the Mamba architecture\cite{mamba}, further integrates MoE with state-space models and employs a gating network to dynamically select informative band combinations, thereby alleviating the information loss caused by static dimensionality reduction. However, existing hyperspectral MoE methods mainly employ expert routing for channel compression or coarse-grained feature distribution. They still treat spectral bands as relatively independent combinations and have not established a unified mechanism that couples dynamic band grouping with the nonlinear topology of continuous bands, namely the spectral manifold. Such band processing lacks sufficient physical structure awareness and limits fine-grained feature allocation in complex scenes. This limitation motivates our exploration of adaptive spectral routing based on the intrinsic physical properties and manifold structure of hyperspectral bands.

\subsection{Prompt Learning}
Prompt learning\cite{PromptTuning}, originally developed in natural language processing, adapts frozen pretrained models to downstream tasks by introducing only a small number of learnable parameters. This parameter-efficient paradigm has since been extended to computer vision through representative methods such as Visual Prompt Tuning (VPT)\cite{VPT} and AdaptFormer\cite{AdaptFormer}, demonstrating that large visual backbones can be efficiently adapted without full fine-tuning. EVP\cite{EVP} further combines prompt learning with high-frequency feature enhancement and quaternion networks\cite{DPLQ} to facilitate cross-modal and domain-specific knowledge transfer.


Prompt-based adaptation has increasingly narrowed modality gaps in visual tracking, from unified representation learning in ProTrack\cite{ProTrack} and modality-specific prompts in ViPT\cite{ViPT} to bidirectional interaction in BAT\cite{BAT} and hyperspectral extensions such as SP-HST\cite{SP-HST}, PHTrack\cite{PHTrack}, and ProFiT\cite{ProFiT}. However, most existing prompts remain static or rely on shallow modality fusion, making them vulnerable to appearance variations and tracking drift in long sequences. To overcome this limitation, we introduce Dual-Stream Spectral-Conditioned Prompt Modulation (DSCPM) with a decoupled condition evolution strategy, enabling prompts to adapt dynamically to evolving spectral context and target appearance without additional parameter updates during inference.

\subsection{Manifold Learning}
In hyperspectral data processing, continuous spectral responses and complex spectral-spatial interactions form a high-dimensional nonlinear space. Manifold learning, such as locally linear embedding (LLE)\cite{LLE, manifold-review}, provides an effective way to uncover its intrinsic low-dimensional structure. RLMR\cite{RLMR} preserves local manifold geometry through hierarchical neighborhood selection and adaptive weighting, while LML\cite{LML} converts local geometric relations into sparse graph structures for semi-supervised classification. To model broader inter-band dependencies, FRPCALG\cite{FRPCALG} combines Laplacian graph regularization with robust principal component analysis, and LLRSC\cite{LLRSC} further integrates low-rank subspace clustering with local manifold constraints. MPWR\cite{MPWR} performs unsupervised band selection using manifold-preserving importance and weak redundancy constraints, whereas Manifold Ranking\cite{ManifoldRanking} formulates salient band selection as a graph-based ranking problem. More recently, MSDiff\cite{MSDiff} has explored diffusion modeling in low-dimensional manifold space to improve robust spectral-spatial representation under degraded conditions.

Although manifold representations provide a valuable foundation for hyperspectral spectral-spatial modeling, the spectral manifold topology may continuously deform and drift in long sequences because of severe illumination changes, occlusion, and target deformation. Such dynamic evolution is difficult to capture in real time using static dimensionality reduction, fixed graph constraints, or non-adaptive interaction mechanisms. Therefore, integrating real-time content awareness with online dynamic graph modeling and adaptive feature routing directly in the original nonlinear spectral manifold space is emerging as a key research direction for advancing hyperspectral tracking toward long-term, dynamic, and robust perception.

\section{Proposed Method}
\begin{figure*}[htbp]
    \centering
    \includegraphics[width=1.0\textwidth]{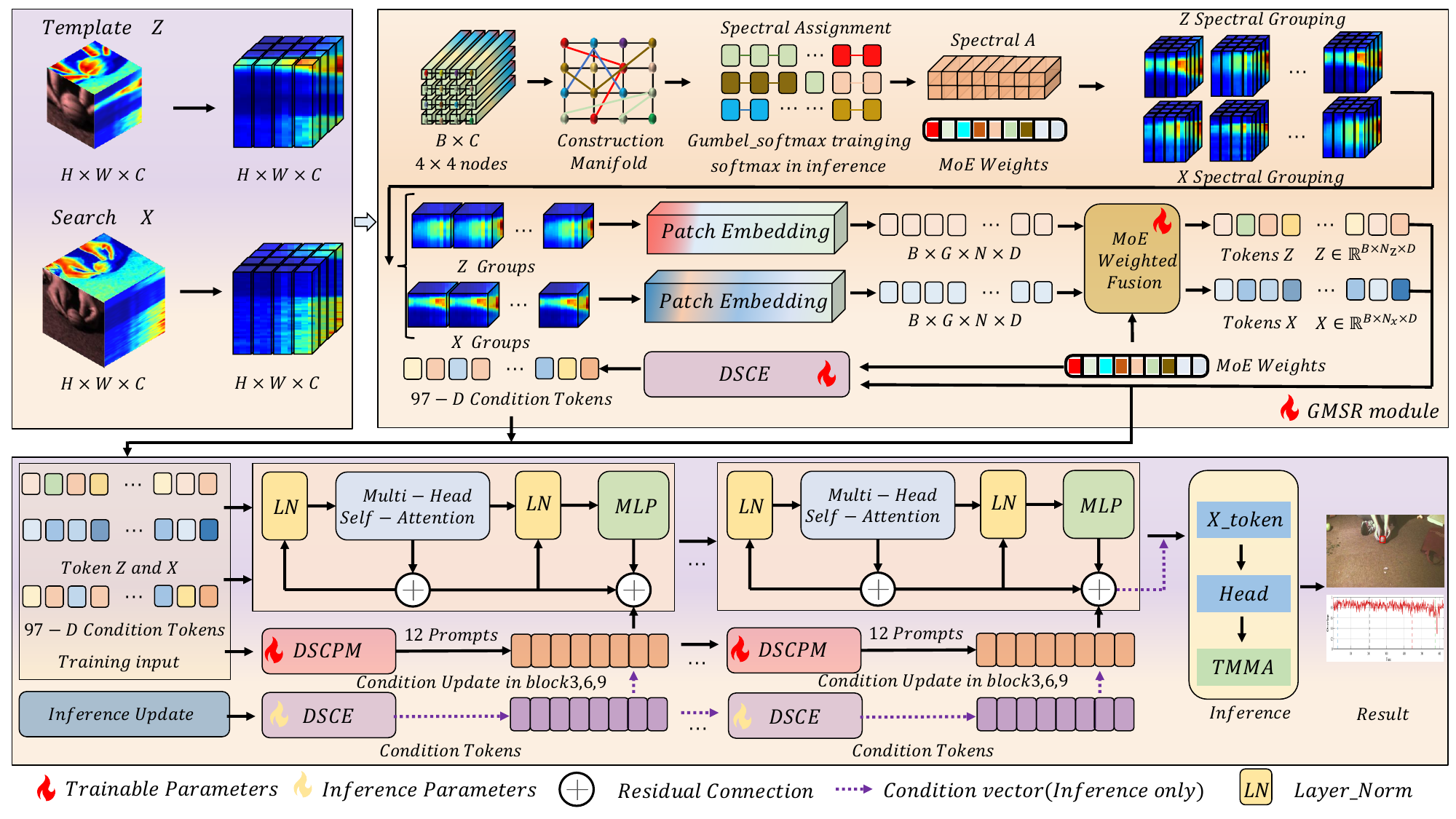}
    \caption{The overall framework diagram of MSP-Net, including the Graph Driven Manifold Spectral Routing (GMSR) module, Dual Stream Spectral Conditional Prompt Modulation (DSCPM) module, Decoupled · Spectral Condition Evolution (DSCE) module, and Temporal Pathology Memory Adaptive Post Processing (TMMA) module.}
    \label{fig:2}
\end{figure*}

This section introduces the proposed Manifold-Guided Spectral Prompt Network (MSP-Net) for hyperspectral object tracking, whose overall data flow is illustrated in Fig.\ref{fig:2}. The network consists of four main components: the Graph-Driven Manifold Spectral Routing (GMSR) module, the Dual-Stream Spectral-Conditioned Prompt Modulation (DSCPM) module, the Decoupled Spectral Condition Evolution (DSCE) strategy, and the Temporal Morphology Memory-Adaptive (TMMA) post-processing module.

\subsection{Graph-Driven Manifold Spectral Routing}

Considering hyperspectral images contain numerous highly redundant bands that form a nonlinear spectral manifold, whereas conventional grouping methods rely on fixed wavelength order and may disrupt intrinsic inter-band relationships, limiting their robustness to appearance and illumination changes. Thus, we propose Graph-Driven Manifold Spectral Routing (GMSR), which formulates spectral grouping as a learnable routing problem. By treating each band as a graph node and dynamically constructing edges according to feature similarity, GMSR enables information to propagate and aggregate across semantically related bands, thereby achieving scene-adaptive spectral grouping beyond fixed wavelength constraints.

In terms of specific implementation, for the input hyperspectral tensor ($X \in \mathbb{R}^{B \times C \times H \times W}$), adaptive average pooling is first applied over the spatial dimensions to obtain the spectral-band node representation (V):

\begin{equation}
V = \operatorname{AdaptiveAvgPool2d}(X)
\in \mathbb{R}^{B \times C \times d}.
\end{equation}
where $d$ denotes the dimensionality of the compressed node features. Based on the obtained spectral-band node representations, we compute the pairwise squared Euclidean distances between nodes and employ a dynamic Gaussian kernel to construct the feature adjacency matrix ($A \in \mathbb{R}^{B \times C \times C}$), thereby characterizing the underlying spectral manifold:
\begin{equation}
A_{i,j}
=
\exp\left(
-\frac{\left\|V_i - V_j\right\|_2^2}
{2\,\zeta(\sigma)^2 + \epsilon}
\right),
\end{equation}
where $V_i$ and $V_j$ denote the node feature vectors of the $i$-th and $j$-th spectral bands, respectively; $\sigma$ is a learnable scalar parameter that controls the bandwidth of the kernel function; $\zeta(\cdot)$ denotes the Softplus activation function, which ensures a positive variance; and $\epsilon$ is a small constant introduced to prevent numerical instability caused by division by zero.

In the spectral manifold, some bands may exhibit high similarity to many other bands, resulting in graph nodes with large degrees, whereas other bands may be relatively independent and thus have smaller degrees. Directly using the unprocessed adjacency matrix $A$ during graph convolutional message passing may cause the aggregated features of high-degree nodes to grow excessively, potentially leading to gradient explosion, while the features of low-degree nodes may be overwhelmed. Therefore, we compute the node degree matrix $D$ and apply symmetric normalization to the adjacency matrix to stabilize gradient propagation:
\begin{equation}
\widetilde{A} = D^{-\frac{1}{2}} A D^{-\frac{1}{2}},
\qquad
D_{i,i} = \sum_{j} A_{i,j}.
\end{equation}
where $D_{i,i}$ denotes the element in the $i$-th row and $i$-th column on the diagonal of the degree matrix $D$, and $A_{i,j}$ represents the connection weight, or similarity, between the $i$-th node (i.e., the $i$-th spectral band) and the $j$-th node in the adjacency matrix $A$ constructed using the Gaussian kernel.

After obtaining the node feature matrix, information is propagated over the spectral manifold. Specifically, the graph convolutional weight matrix $W_{\mathrm{gcn}} \in \mathbb{R}^{d \times d_{\mathrm{gcn}}}$ is employed to aggregate and update the node features, enabling each node to incorporate spectral information from its neighboring nodes on the manifold:
\begin{equation}
H_{\mathrm{new}}
=
\operatorname{GELU}
\left(
\widetilde{A} V W_{\mathrm{gcn}}
\right).
\end{equation}

\begin{equation}
W_{\mathrm{gcn}}^{(t+1)}
=
W_{\mathrm{gcn}}^{(t)}
-
\eta
\frac{\partial \mathcal{L}}
{\partial W_{\mathrm{gcn}}^{(t)}}.
\end{equation}
where $H_{\mathrm{new}} \in \mathbb{R}^{B \times C \times d_{\mathrm{gcn}}}$ denotes the updated manifold node features, and GELU is the nonlinear activation function. In addition, $t$ denotes the current training iteration, $\eta$ represents the learning rate, and $\mathcal{L}$ denotes the loss computed between the predictions produced by the GCN during forward propagation and the corresponding ground-truth labels.

To capture the target motion state in real time, we employ state-aware dynamic routing to maximize the representation of target-relevant information. Since spectral bands within different wavelength ranges exhibit heterogeneous responses, equally fusing all bands may introduce redundant or even conflicting spectral representations. Therefore, manifold-guided band routing is adopted to adaptively group spectrally correlated bands. During training, Gumbel--Softmax with hard assignment produces discrete one-hot routing decisions. Directly retaining such hard assignments during inference may make the routing boundaries overly sensitive and unstable. Therefore, the routing mechanism is switched to softmax-based soft assignment at inference, which provides smoother and more stable routing without introducing additional parameters. The corresponding formulation is given in Eq.6:
\begin{equation}
N =
\begin{cases}
\operatorname{Gumbel\text{-}Softmax}(L,\tau),
& \text{if Training}, \\[4pt]
\operatorname{Softmax}(L),
& \text{if Inference}.
\end{cases}
\end{equation}
where $N$ denotes the routing assignment matrix, and $L$ represents the logits obtained by projecting the updated manifold node features $H_{\mathrm{new}}$ into a predefined grouping space. 

After obtaining the dynamic routing matrix $N$, it is used to perform weighted aggregation of $H_{\mathrm{new}}$, thereby generating a representative feature $F_{\mathrm{group}}$ for each spectral group:
\begin{equation}
F_{\mathrm{group}}
=
\frac{1}{C}
\sum_{c=1}^{C}
N_{i,c,:}
\odot
H_{\mathrm{new},\,i,c,:}.
\end{equation}

Subsequently, to evaluate the contribution of each spectral group to the current tracking scenario, the group features are projected and normalized to generate the expert-mixture weights $W_{\mathrm{moe}}$:
\begin{equation}
W_{\mathrm{moe}}
=
\operatorname{Softmax}
\left(
F_{\mathrm{group}} W_{\mathrm{proj}}
\right).
\end{equation}
where $W_{\mathrm{proj}}$ denotes the learnable linear projection matrix. The resulting weights $W_{\mathrm{moe}}$ are used not only for subsequent weighted feature fusion, but also as important prior information for the Dual-Stream Spectral-Conditioned Prompt Modulation (DSCPM) module, thereby supporting effective feature perception and modulation.

\subsection{Dual-Stream Spectral-Conditioned Prompt-Modulation}
\begin{figure*}[htbp]
    \centering
    \includegraphics[width=1.0\textwidth]{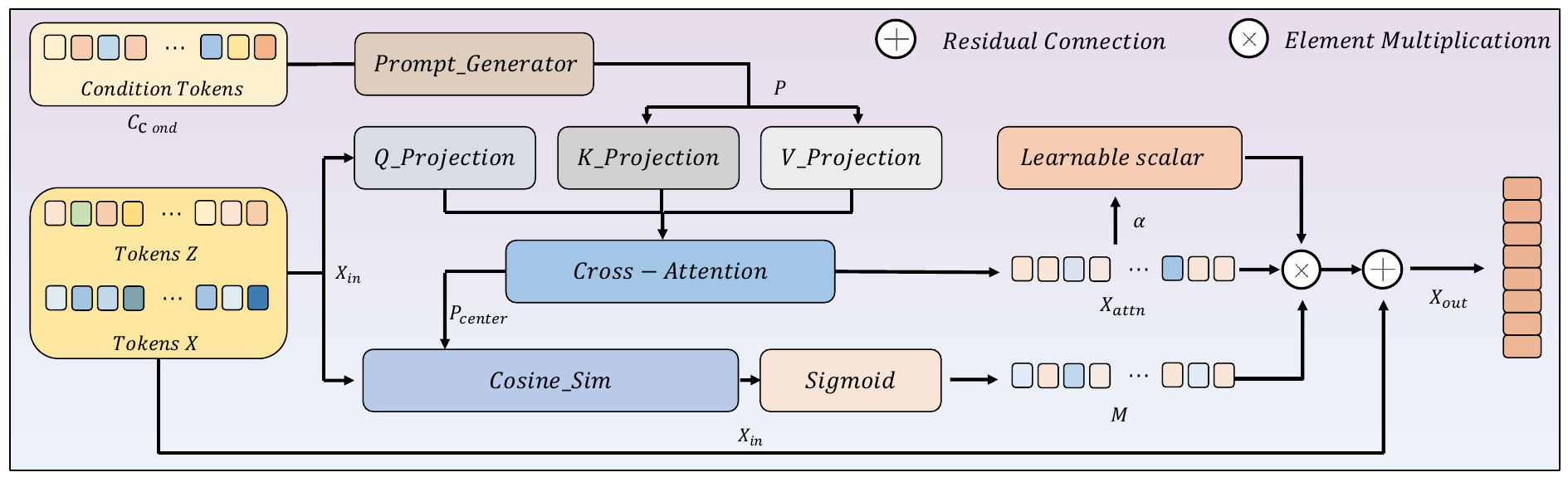}
    \caption{Overview of the DSCPM module. The spectral conditions generated by DSCE are mapped into dynamic prompts, which modulate the joint template–search tokens through cross-attention and cosine gating. A residual connection is then applied to obtain the enhanced feature representation.}
    \label{fig:3}
\end{figure*}

Although GMSR captures manifold-aware inter-band relationships, static prompts remain vulnerable to spectral drift, target deformation, illumination changes, and background interference in hyperspectral videos. To address this limitation, we propose the Dual-Stream Spectral-Conditioned Prompt Modulation (DSCPM) module, as shown in Fig.\ref{fig:3}. DSCPM dynamically generates scene-adaptive prompts by jointly exploiting spectral-manifold context and target appearance cues, enabling the model to distinguish objects with similar spatial structures but different spectral signatures.

Specifically, for each aggregated group feature \(F_{\mathrm{group}}\), we adopt a multidimensional statistical perception strategy to preserve the high-order statistical properties of the spectral manifold in a low-dimensional space. On the one hand, its spatial and channel-wise means are computed to characterize the global low-frequency context layout of the spectral manifold. On the other hand, its standard deviation is calculated to capture high-frequency spectral volatility. Together, these statistics constitute the grouped-band condition vector \(C_{\mathrm{group}}\), which serves as an environmental probe of the manifold space. Meanwhile, the mean and maximum responses of the template features are extracted to generate the template condition vector \(C_{\mathrm{template}}\), which encodes the explicit geometric and textural characteristics of the target. The two heterogeneous information streams are then concatenated with the adaptive mixture weights \(W_{\mathrm{moe}}\) produced by the mixture-of-experts module, yielding the disentangled spectral–target dual-stream condition vector \(C_{\mathrm{cond}}\):
\begin{equation}
C_{\mathrm{cond}}
=
\operatorname{Concat}
\left(
C_{\mathrm{group}},
C_{\mathrm{template}},
W_{\mathrm{moe}}
\right).
\end{equation}

Subsequently, the Prompt Generator takes the condition vector \(C_{\mathrm{cond}}\) as a dynamic reference and maps it end-to-end into a dynamic prompt vector \(P\) tailored to the current scene. To seamlessly inject this dynamic spectral–target prior into the backbone network, the input image tokens \(X_{\mathrm{in}}\) are used as the Query, while the adaptively generated dynamic prompt \(P\) serves as both the Key and Value to perform cross-modal/cross-dimensional cross-attention:
\begin{equation}
X_{\mathrm{attn}}
=
\operatorname{Softmax}
\left(
\frac{
\left(X_{\mathrm{in}} W_q\right)
\left(P W_k\right)^{\top}
}{
\sqrt{d_h}
}
\right)
\left(P W_v\right).
\end{equation}
where \(X_{\mathrm{attn}}\) denotes the attention-modulated features jointly guided by the manifold context and target appearance. It preserves the fine-grained spatial structure of the input image while acquiring enhanced discriminative capability against spectral ambiguity. $W_q$, $W_k$, and $W_v$ are learnable linear projection matrices that map the input features into the query, key, and value spaces, respectively. The variable $d_h$ denotes the feature dimension of each attention head, and $\sqrt{d_h}$ serves as a scaling factor. Since dot-product values can become excessively large in high-dimensional spaces, the Softmax function may be driven into saturated regions with extremely small gradients. Dividing the attention scores by $\sqrt{d_h}$ smooths their distribution, alleviates gradient vanishing, and promotes stable model convergence.

To prevent the prompt information from interfering with background regions, we compute the cosine similarity between $X_{\mathrm{in}}$ and the prompt center, and use the resulting spatial mask $M$ to constrain the modulation process:
\begin{equation}
M
=
\sigma
\left(
\gamma \cdot
\operatorname{CosSim}
\left(
X_{\mathrm{in}},
P_{\mathrm{center}}
\right)
+
\beta
\right).
\end{equation}
where $\sigma$ denotes the Sigmoid function, and $\gamma$ and $\beta$ are learnable parameters. The final modulated feature is given by:
\begin{equation}
X_{\mathrm{out}}
=
X_{\mathrm{in}}
+
\alpha
\left(
M \odot X_{\mathrm{attn}}
\right).
\end{equation}

\subsection{Decoupled Spectral-Condition-Evolution}
\begin{figure*}[htbp]
    \centering
    \includegraphics[width=1.0\textwidth]{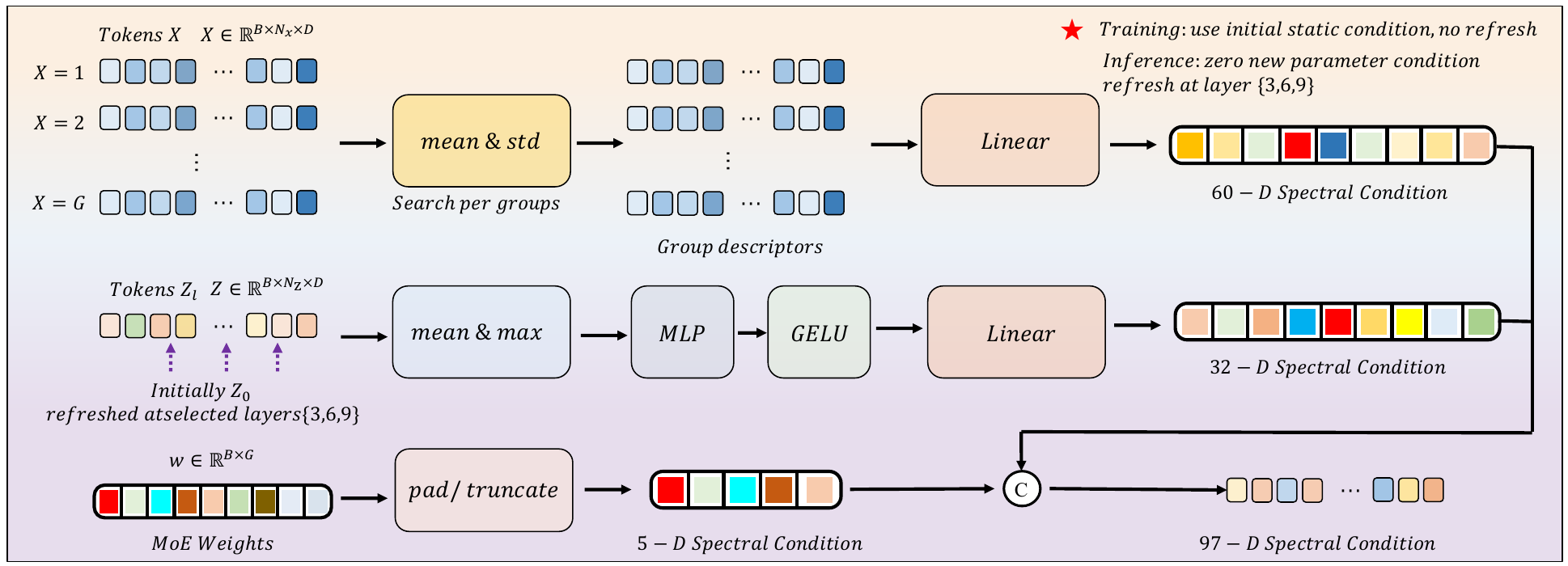}
    \caption{Overview of the DSCE module. The module integrates the grouped search features, current template tokens, and MoE weights to construct a 97-dimensional spectral condition vector. The initial condition is used during training, whereas during inference, the condition is dynamically refreshed at predefined backbone layers using the current template tokens without introducing additional parameters.}
    \label{fig:4}
\end{figure*}

In long-term hyperspectral tracking, illumination changes, occlusions, and target deformation can make first-frame spectral priors increasingly inconsistent with the current target state, causing prompt degradation and tracking drift. To address this issue without costly online fine-tuning or additional memory networks, we propose Decoupled Spectral Condition Evolution (DSCE), as shown in Fig.\ref{fig:4}. DSCE extracts updated target representations from intermediate template tokens at selected Transformer layers and reroutes them to dynamically evolve the spectral conditions. This establishes an online prompt-calibration loop without updating network parameters or introducing additional inference parameters.

In detail, at the $l$-th layer of the visual Transformer, the input consists of the concatenated template and search tokens, denoted as $T^{(l-1)} = [Z^{(l-1)}, X^{(l-1)}]$. After processing by multi-head self-attention and the feed-forward network, the output of the current layer is obtained as:
\begin{equation}
T^{(l)}
=
\operatorname{Block}_{l}
\left(
T^{(l-1)}
\right).
\end{equation}

To capture the target’s latest physical state in the current frame without introducing additional parameters, we dynamically split and decouple the updated joint sequence along the sequence dimension. Specifically, the first \(I_z\) tokens are extracted as the deep template representation \(Z^{(l)}\), which encodes rich temporal and environmental context at the current network depth:
\begin{equation}
Z^{(l)}
=
T^{(l)}_{[:,\,0:I_z]}.
\end{equation}
Based on the extracted deep template features $Z^{(l)}$, the DSCE module updates the spectral condition vector $C_{\mathrm{cond}}$ through a discrete gated condition-evolution mechanism. To achieve the optimal Pareto balance between computational efficiency and temporal update frequency, we define a set of update-layer indices $\Omega$ within the backbone network. In our implementation, $\Omega={3,6,9}$. The layer-wise evolution rule of the spectral condition is formulated as follows:
\begin{equation}
C_{\mathrm{cond}}^{(l)}
=
\begin{cases}
\mathcal{X}_{\mathrm{ext}}
\left(
F_{\mathrm{group}},
W_{\mathrm{moe}},
Z^{(l)}
\right),
& l \in \Omega \ \text{inference}, \\[4pt]
C_{\mathrm{cond}}^{(l-1)},
& \text{otherwise}.
\end{cases}
\end{equation}
where $\mathcal{X}_{\mathrm{ext}}(\cdot)$ denotes the dual-stream condition extractor defined in DSCPM, $\mathcal{F}_{\mathrm{group}}$ represents the grouped and modulated spectral features, and $W_{\mathrm{moe}}$ denotes the corresponding mixture-of-experts weights.

Subsequently, for prompt modulation at the $l$-th layer, the prompt generator produces the corresponding visual prompt based on the latest spectral condition:
\begin{equation}
P^{(l)}
=
\mathcal{G}_{\mathrm{prompt}}
\left(
C_{\mathrm{cond}}^{(l)}
\right).
\end{equation}
where $P^{(l)}$ denotes the sequence of dynamic visual prompt vectors generated at the $l$-th layer of the backbone. It encodes contextual guidance specific to the feature space at the current depth and is subsequently used directly as the key and value in the cross-attention modulation of DSCPM. $\mathcal{G}_{\mathrm{prompt}}$ denotes the mapping function of the prompt generator network, while $C_{\mathrm{cond}}^{(l)}$ represents the latest spectral condition vector obtained through the parameter-free evolution of the DSCE module at the $l$-th layer.

During training, i.e., $\mathrm{Phase}=\mathrm{Training}$, the condition evolution mechanism remains inactive, such that $C_{\mathrm{cond}}^{(l)}=C_{\mathrm{cond}}^{(0)}$. The model parameters are updated solely through global gradient optimization, ensuring stable convergence of the backbone network. During inference, the DSCE module is dynamically activated, progressively recalibrating the visual prompts $P^{(l)}$ as the network depth increases, thereby enabling adaptive responses to substantial target appearance variations. By decoupling condition evolution from parameter optimization, this strategy achieves dynamic state adaptation without introducing additional parameters or requiring online fine-tuning. Consequently, it alleviates the feature mismatch caused by conventional static prompts in long-term tracking and improves long-term tracking robustness.

\subsection{ Temporal Morphology Memory-Adaptive Post-processing}

\begin{figure}[!t]
\centering
\includegraphics[width=3.5in]{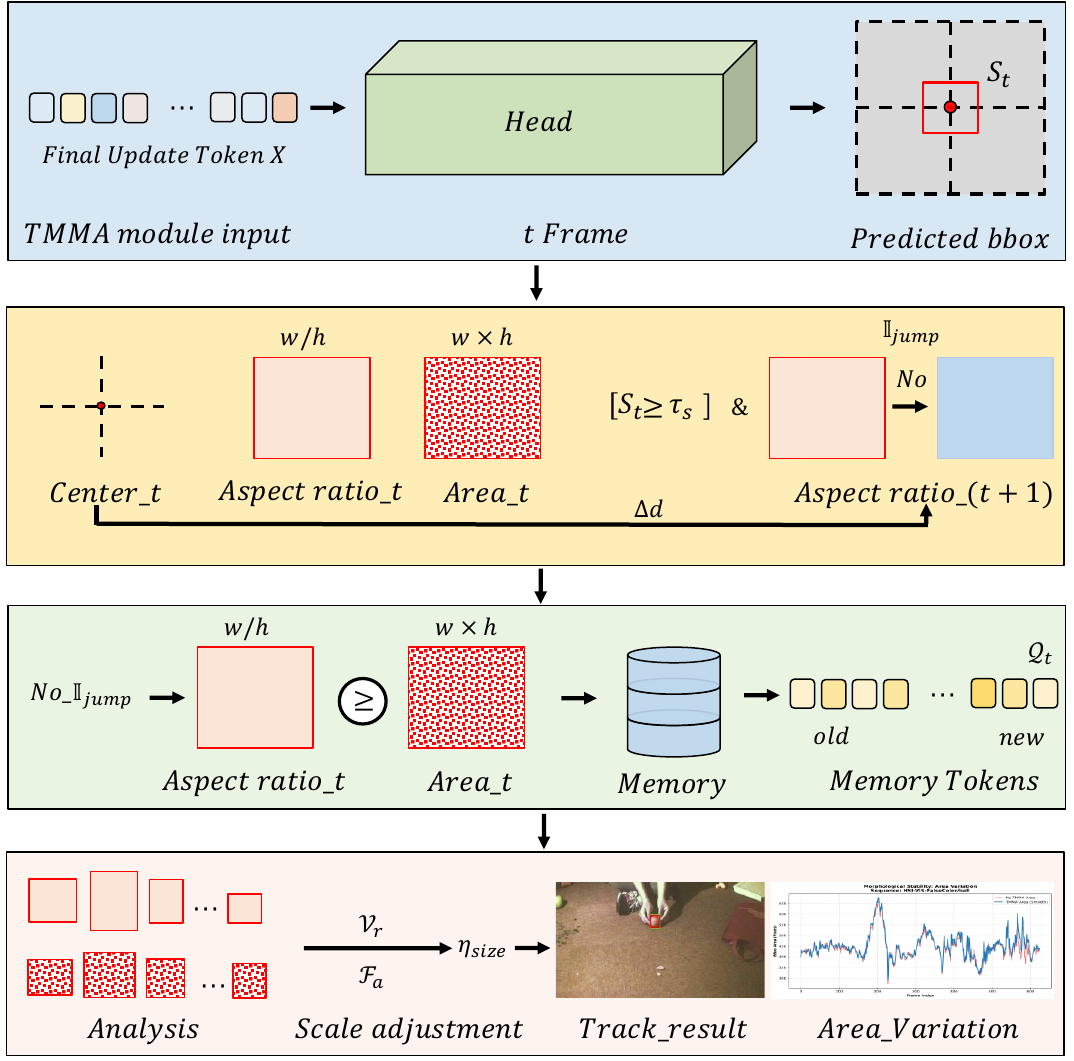}
\caption{Framework of the TMMA module. The module constructs a temporal morphology memory from historical tracking states and adaptively adjusts the search region and scale parameters according to the current predicted bounding box and confidence score, thereby improving tracking stability and adaptability in long sequences.}
\label{fig:5}
\end{figure}

Conventional post-processing relies on static scale penalties and Hann windows to smooth predictions, but rigid constraints may suppress genuine target deformation, whereas relaxed constraints increase drift toward spectrally similar distractors. To balance stability and adaptability, we propose Temporal Morphological Memory Adaptation (TMMA), as shown in Fig.\ref{fig:5}. TMMA compares the predicted spatial displacement and morphological variation with reliable historical states through Spatial Jump Lock and Morphological Fluctuation Sensing, allowing it to distinguish genuine non-rigid deformation from distractor-induced shifts and adaptively stabilize the bounding box.

Firstly, let the initial bounding box predicted by the network at frame $t$ be denoted as $B_t=(x_t,y_t,w_t,h_t)$, with its center coordinates given by $C_t=(x_t+w_t/2,y_t+h_t/2)$. To prevent abnormal spatial position mutations in the target bounding box induced by similar interferences, this paper constructs an adaptive spatial mutation lock threshold based on the spatial scale of the target itself. Specifically, the Euclidean distance $\Delta d$ between the current center $C_t$ and the center of the previous frame $C_{t-1}$, together with the diagonal length $D_t$ of the current target, is computed as follows:
\begin{equation}
\Delta d
=
\left\|
C_t-C_{t-1}
\right\|_2,
\qquad
D_t
=
\sqrt{
w_t^2+h_t^2
}.
\end{equation}

Accordingly, the spatial-jump indicator function $\mathbb{I}_{\mathrm{jump}}$ is defined as:
\begin{equation}
\mathbb{I}_{\mathrm{jump}}
=
\begin{cases}
1, & \Delta d > \lambda D_t, \\[4pt]
0, & \text{otherwise}.
\end{cases}
\end{equation}
where $\lambda$ denotes the relative displacement tolerance, which is set to a stricter value of $\lambda=0.5$ in our implementation. When $\mathbb{I}_{\mathrm{jump}}=1$, the system identifies the current prediction as an abrupt jitter caused by a similar distractor and rejects the corresponding frame state from being stored in the memory bank.

Provided that the tracking confidence score $s_t$ exceeds the reliability threshold $\tau_s$ and no spatial jump is detected, i.e., $\mathbb{I}_{\mathrm{jump}}=0$, we extract the target morphology features, including the aspect ratio $r_t$ and area $a_t$, to construct the temporal morphology memory bank:
\begin{equation}
r_t
=
\frac{w_t}{h_t+\epsilon},
\qquad
a_t
=
w_t h_t.
\end{equation}
The resulting morphology tuple is then stored in a sliding-window memory bank $\mathcal{Q}_t$ of length $K$, where $K=8$ in our implementation. This memory bank retains only historical target morphology states associated with high-confidence predictions and stable motion. To determine whether the target is undergoing sustained and substantial deformation, we compute recent morphology statistics from the memory bank. Specifically, we calculate the variance $\mathcal{V}_r$ of the aspect-ratio sequence and the max-to-min fluctuation rate $\mathcal{F}_a$ of the area sequence:
\begin{equation}
\mathcal{V}_r
=
\frac{1}{\left|\mathcal{Q}_t\right|}
\sum_{i\in\mathcal{Q}_t}
\left(
r_i-\bar{r}
\right)^2.
\end{equation}

\begin{equation}
\mathcal{F}_a
=
\frac{
\max\left(\mathcal{Q}_t(a)\right)
}{
\min\left(\mathcal{Q}_t(a)\right)+\epsilon
}
-1.
\end{equation}
where $\bar{r}$ denotes the mean aspect ratio stored in the memory bank, and $\epsilon$ is a small constant introduced to prevent division-by-zero overflow. Based on the above morphology statistics, the TMMA module performs discrete gated modulation of the scale penalty factor $\eta_{\mathrm{size}}$ for the current frame:
\begin{equation}
\eta_{\mathrm{size}}
=
\begin{cases}
1.00,
& \mathcal{V}_r>\mu_r
\ \text{or}\
\mathcal{F}_a>\mu_a, \\[4pt]
\eta_{\mathrm{base}},
& \text{otherwise}.
\end{cases}
\end{equation}
where $\mu_r$ and $\mu_a$ denote the variance threshold and area-fluctuation threshold used to trigger deformation awareness, respectively. In our implementation, they are set to $\mu_r=0.04$ and $\mu_a=0.20$. $\eta_{\mathrm{base}}$ denotes the baseline penalty for background suppression and is set to $0.99$.

\section{Experiments}
\subsection{ Experiments Setup}
All experiments were implemented in PyTorch on an Intel Core i7-14700 CPU with two NVIDIA RTX 3090 GPUs. AdamW was used with an initial learning rate and weight decay of $1\times10^{-4}$, a batch size of 32, and 30 training epochs. The search and template images were resized to $256\times256$ and $128\times128$ pixels, respectively. In addition, to effectively evaluate MSPNet, the model is trained exclusively on the HOT2024 training set and tested on the validation sets of HOT2020\cite{MHT}, HOT2023(https://www.hsitracking.com/), and HOT2024(https://www.hsitracking.com/). Considering the distributional differences, IMEC25\cite{IMEC25} is trained and evaluated separately.

\begin{table}[t]
    \centering
    \caption{Performance comparison in terms of computational complexity.}
    \label{tab:complexity_comparison}

    \begingroup
    \fontsize{7}{8}\selectfont
    \setlength{\tabcolsep}{3.0pt}
    \renewcommand{\arraystretch}{1.10}

    \begin{tabular*}{\columnwidth}
    {@{\extracolsep{\fill}}lccc@{}}
        \toprule
        \textbf{Method}
        & \textbf{FPS}
        & \textbf{FLOPs (G)}
        & \textbf{Device} \\
        \midrule

        DeepHKCF\cite{DeepHKCF}
        & 0.91
        & --
        & CPU \\

        CNHT\cite{CNHT}
        & 2.61
        & --
        & CPU \\

        MHT\cite{MHT}
        & 2.23
        & --
        & CPU \\

        BAE-Net\cite{BAE-Net}
        & 0.72
        & --
        & CPU \\

        SiamBAG\cite{SiamBAG}
        & 10.5
        & 126.3
        & GPU \\

        SEE-Net\cite{SEE-Net}
        & 12.8
        & 244.2
        & GPU \\

        MMF-Net\cite{MMF-Net}
        & 6.1
        & 352.6
        & GPU \\

        SSTrack\cite{SSTRACK}
        & 21.62
        & 135.79
        & GPU \\

        SPIRIT\cite{SPIRIT}
        & 26.2
        & 133.4
        & GPU \\

        Trans-DAT\cite{Trans-DAT}
        & 23.2
        & \textcolor{red}{\textbf{30.3}}
        & GPU \\

        TIPSST\cite{TIPSST}
        & 26.6
        & 136.5
        & GPU \\

        HOT-MoE\cite{HOT-MOE}
        & \textcolor{blue}{\textbf{30.2}}
        & \textcolor{blue}{\textbf{58.37}}
        & GPU \\

        DRSST-Net\cite{DRSST}
        & 22.3
        & 62.52
        & GPU \\

        \textbf{MSP-Net}
        & \textcolor{red}{\textbf{34.71}}
        & 73.01
        & GPU \\

        \bottomrule
    \end{tabular*}

    \vspace{2pt}

    \noindent
    \parbox{\columnwidth}{%
        \centering
        \fontsize{7}{8}\selectfont
        Note: The best and second-best results are highlighted in
        red and blue, respectively.
    }

    \endgroup
\end{table}
As shown in Table~\ref{tab:complexity_comparison}, MSP-Net achieves the highest inference speed of 34.71 FPS with a moderate computational cost of 73.01 GFLOPs. Although its FLOPs are slightly higher than those of Trans-DAT and HOT-MoE, MSP-Net provides a better balance between tracking accuracy and efficiency, demonstrating strong practical potential for real-time hyperspectral object tracking.

\subsection{Comparison on HOT2020 Dataset}
As shown in Table~\ref{tab:hot2020_comparison}, the proposed MSP-Net achieves outstanding tracking performance on the HOT2020 hyperspectral evaluation dataset, attaining a success score (AUC) of 0.8072 and a precision score (DP@20) of 0.9756. Compared with several mainstream and advanced tracking methods developed in recent years, MSP-Net demonstrates clear performance advantages. In particular, compared with the current state-of-the-art ProFit model, which achieves an AUC of 0.7580 and a DP@20 score of 0.9710, MSP-Net improves the success score and precision score by 4.92\% and 0.46 \%, respectively. These improvements are further illustrated intuitively in Fig.\ref{fig:6}.
\begin{table}[t]
    \centering
    \caption{Comparison with state-of-the-art hyperspectral trackers on the HOT2020 benchmark.}
    \label{tab:hot2020_comparison}

    \begingroup
    \fontsize{7}{8}\selectfont
    \setlength{\tabcolsep}{3.0pt}
    \renewcommand{\arraystretch}{1.08}

    \begin{tabular*}{\columnwidth}
    {@{\extracolsep{\fill}}lccc@{}}
        \toprule
        \textbf{Algorithm}
        & \textbf{Journal/Year}
        & \textbf{AUC}
        & \textbf{DP@20} \\
        \midrule

        BAE-Net\cite{BAE-Net}
        & ICIP/2020
        & 0.6158
        & 0.8766 \\

        SEE-Net\cite{SEE-Net}
        & TIP/2023
        & 0.6752
        & 0.9321 \\

        SiamBAG\cite{SiamBAG}
        & TGRS/2023
        & 0.6320
        & 0.8930 \\

        SSTrack\cite{SSTRACK}
        & IF/2024
        & 0.7228
        & 0.9551 \\

        SENSE\cite{SENSE}
        & IF/2024
        & 0.6691
        & 0.9502 \\

        MMF-Net\cite{MMF-Net}
        & TGRS/2024
        & 0.7002
        & 0.9308 \\

        SPIRIT\cite{SPIRIT}
        & TGRS/2024
        & 0.6884
        & 0.9237 \\

        PHTrack\cite{PHTrack}
        & TGRS/2024
        & 0.6699
        & 0.9180 \\

        DASSP\cite{DASSP}
        & PR/2025
        & 0.6914
        & 0.9163 \\

        HOT-MoE\cite{HOT-MOE}
        & TMM/2025
        & 0.7223
        & 0.9418 \\

        HyMamba\cite{HyMamba}
        & TIP/2025
        & 0.7300
        & 0.9630 \\

        ProFit\cite{ProFiT}
        & ISPRS/2025
        & \textcolor{blue}{\textbf{0.7580}}
        & \textcolor{blue}{\textbf{0.9710}} \\

        TIPSST\cite{TIPSST}
        & ASC/2026
        & 0.6870
        & 0.8830 \\

        CSSTrack\cite{CSSTrack}
        & PR/2026
        & 0.7030
        & 0.9510 \\

        DRSST-Net\cite{DRSST}
        & ISPRS/2026
        & 0.7200
        & 0.9420 \\

        \midrule

        \textbf{MSP-Net}
        & --
        & \textcolor{red}{\textbf{0.8072}}
        & \textcolor{red}{\textbf{0.9756}} \\

        \bottomrule
    \end{tabular*}

    \vspace{2pt}

    \noindent
    \parbox{\columnwidth}{%
        \centering
        \fontsize{7}{8}\selectfont
        Note: Red and blue denote the best and second-best results, respectively.
    }

    \endgroup
\end{table}

\begin{figure}[htbp]
    \centering
    \begin{minipage}{0.24\textwidth}
        \centering
        \includegraphics[width=\linewidth]{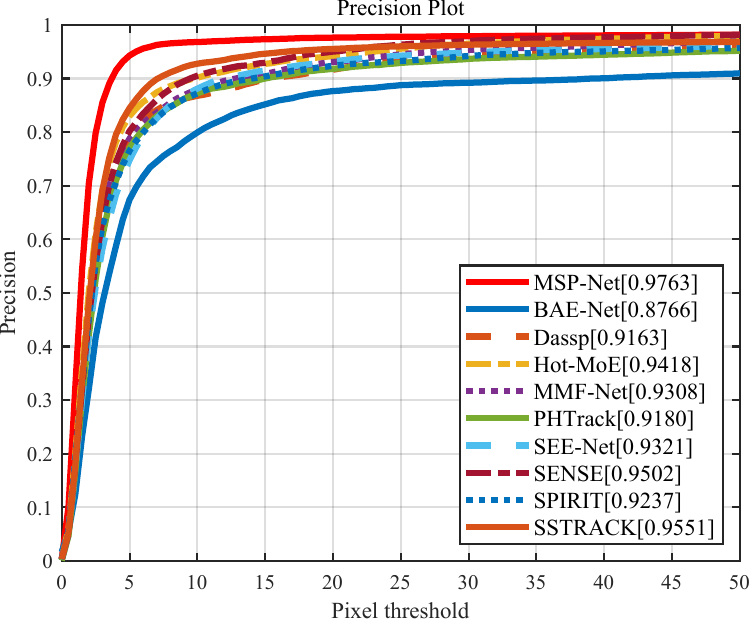}
        \label{fig:6}
    \end{minipage}\hfill
    \begin{minipage}{0.24\textwidth}
        \centering
        \includegraphics[width=\linewidth]{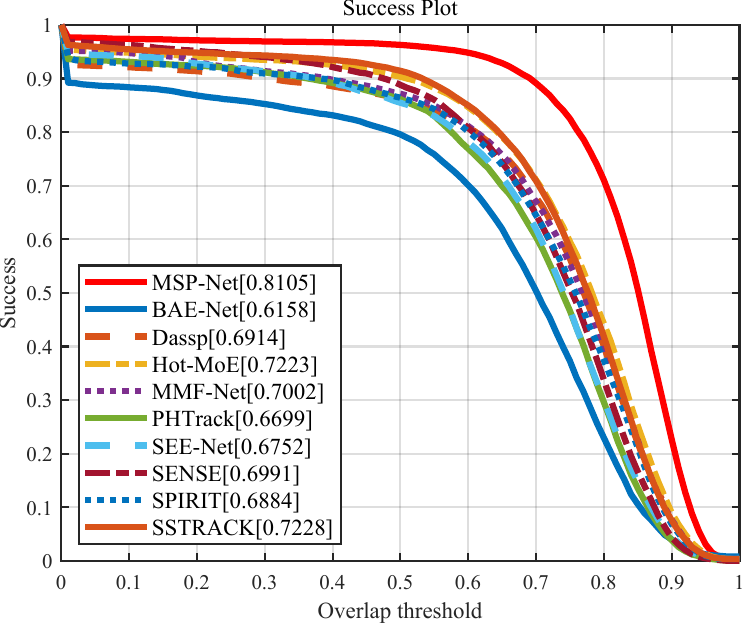}
        \label{fig:6}
    \end{minipage}
    \caption{Precision and success plots of all competing hyperspectral trackers on the HOT2020 dataset.}   
    \label{fig:6}
\end{figure}

To further evaluate the applicability of MSP-Net across common tracking scenarios, we conduct an attribute-based evaluation on HOT2020 under 11 challenging conditions: fast motion (FM), background clutter (BC), illumination variation (IV), out-of-plane rotation (OPR), occlusion (OCC), motion blur (MB), low resolution (LR), scale variation (SV), in-plane rotation (IPR), out-of-view motion (OV), and deformation (DEF). The DP@20 metric is employed to assess the tracking performance of MSP-Net under each challenging attribute.
\begin{table*}[t]
    \centering
    \caption{Attribute-based comparison on the HOT2020 dataset.
    The top-performing values are highlighted in red and blue.}
    \label{tab:attribute_hot2020}

    \begingroup
    \fontsize{7}{8}\selectfont
    \setlength{\tabcolsep}{3.0pt}
    \renewcommand{\arraystretch}{1.08}

    \begin{tabular*}{\textwidth}
    {@{\extracolsep{\fill}}lccccccccccc@{}}
        \toprule
        \textbf{Tracker}
        & \textbf{BC}
        & \textbf{DEF}
        & \textbf{FM}
        & \textbf{IV}
        & \textbf{IPR}
        & \textbf{LR}
        & \textbf{MB}
        & \textbf{OCC}
        & \textbf{OPR}
        & \textbf{OV}
        & \textbf{SV} \\
        \midrule

        BAE-Net\cite{BAE-Net}
        & 0.921 & 0.940 & 0.871 & 0.816 & 0.985
        & 0.734 & 0.882 & 0.790 & 0.982 & 0.864 & 0.889 \\

        SEE-Net\cite{SEE-Net}
        & 0.959 & 0.938 & 0.991 & 0.874 & 0.983
        & 0.940 & 0.981 & 0.885 & 0.976 & 0.860 & 0.927 \\

        SiamBAG\cite{SiamBAG}
        & 0.899 & 0.936 & 0.883 & 0.930 & 0.846
        & 0.839 & 0.899 & 0.831 & 0.906 & 0.891 & 0.892 \\

        SSTrack\cite{SSTRACK}
        & 0.968 & 0.990 & 0.995 & 0.928 & 0.981
        & 0.944 & 0.989 & 0.928 & 0.989 & 0.986 & 0.947 \\

        SENSE\cite{SENSE}
        & 0.966 & 0.948 & 0.992 & 0.917 & 0.976
        & 0.942 & 0.993 & 0.918 & 0.971 & 0.860 & 0.943 \\

        MMF-Net\cite{MMF-Net}
        & 0.940 & 0.935 & 0.995 & 0.896 & 0.972
        & 0.946 & 0.996 & 0.888 & 0.964 & 0.860 & 0.928 \\

        SPIRIT\cite{SPIRIT}
        & 0.963 & 0.925 & 0.987 & 0.845 & 0.986
        & 0.893 & 0.968 & 0.879 & 0.977 & 0.860 & 0.915 \\

        PHTrack\cite{PHTrack}
        & 0.925 & 0.919 & 0.993 & 0.863 & 0.952
        & 0.891 & 0.986 & 0.880 & 0.947 & 0.855 & 0.914 \\

        DASSP\cite{DASSP}
        & 0.890 & 0.972 & 0.958 & 0.917 & 0.967
        & 0.944 & 0.961 & 0.917 & 0.958 & 0.991 & 0.922 \\

        HOT-MoE\cite{HOT-MOE}
        & 0.929
        & 0.955
        & 0.965
        & \textcolor{blue}{\textbf{0.967}}
        & 0.967
        & \textcolor{red}{\textbf{0.994}}
        & 0.945
        & 0.937
        & 0.961
        & \textcolor{red}{\textbf{1.000}}
        & 0.940 \\

        ProFit\cite{ProFiT}
        & 0.986
        & 0.991
        & \textcolor{red}{\textbf{1.000}}
        & \textcolor{red}{\textbf{0.994}}
        & 0.959
        & 0.973
        & 0.991
        & 0.954
        & 0.989
        & \textcolor{red}{\textbf{1.000}}
        & 0.960 \\

        \midrule

        \textbf{MSP-Net}
        & \textcolor{red}{\textbf{0.995}}
        & \textcolor{red}{\textbf{1.000}}
        & \textcolor{red}{\textbf{1.000}}
        & 0.961
        & \textcolor{red}{\textbf{0.997}}
        & \textcolor{blue}{\textbf{0.984}}
        & \textcolor{red}{\textbf{1.000}}
        & \textcolor{red}{\textbf{0.960}}
        & \textcolor{red}{\textbf{0.995}}
        & \textcolor{red}{\textbf{1.000}}
        & \textcolor{red}{\textbf{0.964}} \\

        \bottomrule
    \end{tabular*}

    \vspace{2pt}

    \noindent
    \parbox{\textwidth}{%
        \centering
        \fontsize{7}{8}\selectfont
        Note: Red and blue denote the best and second-best results, respectively.
    }

    \endgroup
\end{table*}

\begin{figure}[!t]
\centering
\includegraphics[width=3.5in]{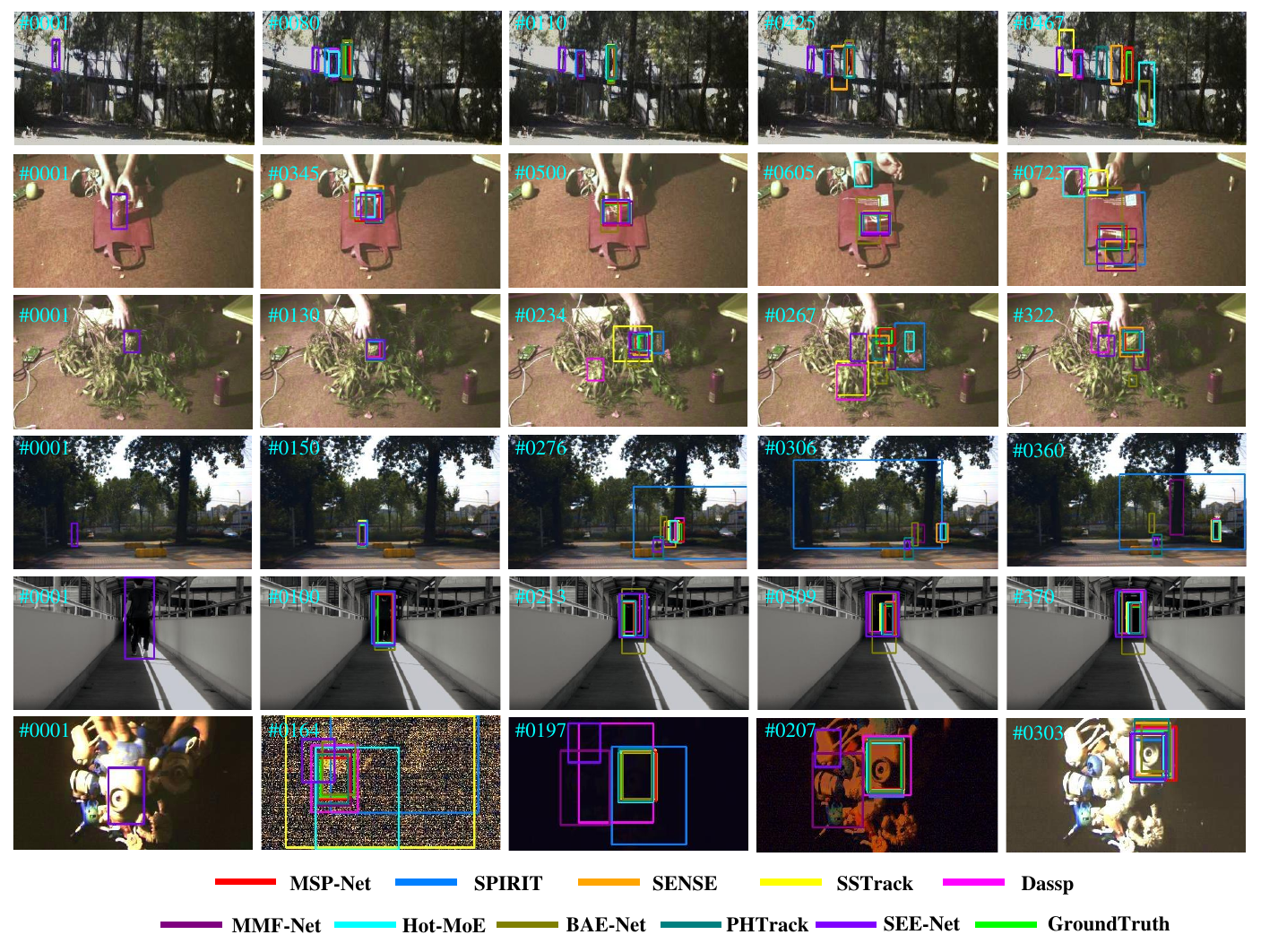}
\caption{Visual comparisons, arranged from top to bottom, cover a set of scenes from the HOT2020 campus: coke, fruit, student, pedestrian2, student, and toy2 scene.}
\label{fig:7}
\end{figure}

As shown in Table~\ref{tab:attribute_hot2020} and Fig.\ref{fig:7}, MSP-Net ranks among the top three methods across all 11 challenging attributes in terms of the DP@20 metric. Specifically, it achieves the best performance in nine scenarios, including background clutter (BC), deformation (DEF), fast motion (FM), in-plane rotation (IPR), motion blur (MB), occlusion (OCC), out-of-plane rotation (OPR), out-of-view motion (OV), and scale variation (SV). It further ranks second under low resolution (LR) and third under illumination variation (IV). These results demonstrate the stable and robust tracking performance of MSP-Net across diverse challenging scenarios.

\subsection{Comparison on HOT2023 Dataset}
In HOT2023, we test MSP-Net on the VIS, RedNIR, and NIR modalities of HOT2023. As shown in Table~\ref{tab:hot2023_comparison}, MSP-Net achieves state-of-the-art AUC and DP@20 performance across the three modalities. It surpasses the previous best methods by 6.52\% in AUC and 0.46\% in DP@20 under NIR, and outperforms ProFit by 8.21\% and 7.89\% under RedNIR. Under VIS, although its DP@20 of 0.9138 is slightly below ProFit’s 0.9150, MSP-Net achieves the highest AUC of 0.7576. These results demonstrate its strong accuracy and robustness across diverse spectral representations and complex scenarios.

\begin{table}[t]
    \centering
    \caption{Comparison with state-of-the-art hyperspectral trackers on the HOT2023 benchmark.}
    \label{tab:hot2023_comparison}

    \renewcommand{\arraystretch}{1.12}
    \setlength{\tabcolsep}{2.2pt}

    \resizebox{\columnwidth}{!}{%
    \begin{tabular}{lccccccc}
        \toprule
        \multirow{2}{*}{\textbf{Tracker}}
        & \multirow{2}{*}{\textbf{Venue/Year}}
        & \multicolumn{2}{c}{\textbf{VIS2023}}
        & \multicolumn{2}{c}{\textbf{NIR2023}}
        & \multicolumn{2}{c}{\textbf{RedNIR2023}} \\
        \cmidrule(lr){3-4}
        \cmidrule(lr){5-6}
        \cmidrule(lr){7-8}
        & & \textbf{AUC} & \textbf{DP@20}
        & \textbf{AUC} & \textbf{DP@20}
        & \textbf{AUC} & \textbf{DP@20} \\
        \midrule

        BAE-Net\cite{BAE-Net}
        & ICIP/2020
        & 0.5798 & 0.8199
        & 0.4515 & 0.7805
        & 0.4000 & 0.5191 \\

        VIPT\cite{ViPT}
        & CVPR/2023
        & 0.6501 & 0.8577
        & 0.7355 & 0.9300
        & 0.5735 & 0.7041 \\

        Trans-DAT\cite{Trans-DAT}
        & TCSVT/2023
        & 0.4932 & 0.6594
        & 0.6728 & 0.8822
        & 0.5442 & 0.7150 \\

        SiamBAG\cite{SiamBAG}
        & TGRS/2023
        & 0.5630 & 0.7920
        & 0.4790 & 0.7610
        & 0.2890 & 0.4110 \\

        SSTrack\cite{SSTRACK}
        & Inf. Fusion/2024
        & 0.6669 & 0.8719
        & 0.6601 & 0.8485
        & 0.4085 & 0.4988 \\

        SENSE\cite{SENSE}
        & Inf. Fusion/2024
        & 0.6170 & 0.8260
        & 0.5529 & 0.7627
        & 0.4021 & 0.5001 \\

        SPIRIT\cite{SPIRIT}
        & TGRS/2024
        & 0.6175 & 0.8203
        & 0.6227 & 0.8261
        & 0.3891 & 0.5007 \\

        PHTrack\cite{PHTrack}
        & TGRS/2024
        & 0.5900 & 0.7949
        & 0.5410 & 0.7654
        & 0.4226 & 0.5113 \\

        DASSP\cite{DASSP}
        & PR/2025
        & 0.6593 & 0.8620
        & 0.7379 & 0.9324
        & 0.5707 & 0.7038 \\

        HOT-MoE\cite{HOT-MOE}
        & TMM/2025
        & 0.6625 & 0.8547
        & 0.7158 & 0.9148
        & 0.5114 & 0.6334 \\

        HyMamba\cite{HyMamba}
        & TIP/2025
        & 0.6800 & 0.8960
        & 0.7530
        & \textcolor{blue}{\textbf{0.9620}}
        & 0.5430 & 0.6830 \\

        ProFit\cite{ProFiT}
        & ISPRS/2025
        & \textcolor{blue}{\textbf{0.7200}}
        & \textcolor{red}{\textbf{0.9150}}
        & \textcolor{blue}{\textbf{0.7540}}
        & 0.9470
        & \textcolor{blue}{\textbf{0.6130}}
        & \textcolor{blue}{\textbf{0.7550}} \\

        \midrule

        \textbf{MSP-Net}
        & --
        & \textcolor{red}{\textbf{0.7576}}
        & \textcolor{blue}{\textbf{0.9138}}
        & \textcolor{red}{\textbf{0.8192}}
        & \textcolor{red}{\textbf{0.9666}}
        & \textcolor{red}{\textbf{0.6951}}
        & \textcolor{red}{\textbf{0.8339}} \\

        \bottomrule
    \end{tabular}%
    }

    \vspace{2pt}

    \parbox{\columnwidth}{%
        \centering
        \scriptsize
        Note: Red and blue indicate the best and second-best results, respectively.
    }
\end{table}

\begin{table}[t]
    \centering
    \caption{Comparison with state-of-the-art hyperspectral trackers on the HOT2024 benchmark.}
    \label{tab:hot2024_comparison}

    \setlength{\tabcolsep}{2.0pt}
    \renewcommand{\arraystretch}{1.12}

    \resizebox{\columnwidth}{!}{%
    \begin{tabular}{lccccccc}
        \toprule
        \multicolumn{2}{c}{}
        & \multicolumn{2}{c}{\textbf{VIS2024}}
        & \multicolumn{2}{c}{\textbf{NIR2024}}
        & \multicolumn{2}{c}{\textbf{RedNIR2024}} \\
        \cmidrule(lr){3-4}
        \cmidrule(lr){5-6}
        \cmidrule(lr){7-8}

        \textbf{Algorithm}
        & \textbf{Journal/Year}
        & \textbf{AUC}
        & \textbf{DP@20}
        & \textbf{AUC}
        & \textbf{DP@20}
        & \textbf{AUC}
        & \textbf{DP@20} \\
        \midrule

        BAE-Net\cite{BAE-Net}
        & ICIP/2020
        & 0.3290 & 0.4900
        & 0.4630 & 0.7670
        & 0.3070 & 0.4290 \\

        SiamBAG\cite{SiamBAG}
        & TGRS/2023
        & 0.3970 & 0.5590
        & 0.5360 & 0.7830
        & 0.2260 & 0.3580 \\

        Trans-DAT\cite{Trans-DAT}
        & TCSVT/2023
        & 0.4173 & 0.5303
        & 0.6076 & 0.7498
        & 0.4358 & 0.5473 \\

        SSTrack\cite{SSTRACK}
        & Inf. Fusion/2024
        & 0.3994 & 0.4883
        & 0.7235 & 0.8851
        & \textcolor{red}{\textbf{0.5205}}
        & \textcolor{red}{\textbf{0.6589}} \\

        SENSE\cite{SENSE}
        & Inf. Fusion/2024
        & 0.3094 & 0.3991
        & 0.5878 & 0.7847
        & 0.3748 & 0.4651 \\

        MMF-Net\cite{MMF-Net}
        & TGRS/2024
        & 0.4880 & 0.6450
        & 0.7010 & 0.8750
        & 0.3950 & 0.5210 \\

        SPIRIT\cite{SPIRIT}
        & TGRS/2024
        & 0.3306 & 0.4082
        & 0.6842 & 0.8427
        & 0.3874 & 0.5158 \\

        PHTrack\cite{PHTrack}
        & TGRS/2024
        & 0.3151 & 0.4123
        & 0.5517 & 0.7495
        & 0.2722 & 0.3644 \\

        UBSTrack\cite{UBSTrack}
        & TGRS/2025
        & \textcolor{red}{\textbf{0.5474}}
        & \textcolor{blue}{\textbf{0.6873}}
        & \textcolor{blue}{\textbf{0.7575}}
        & \textcolor{blue}{\textbf{0.8856}}
        & \textcolor{blue}{\textbf{0.5047}}
        & 0.6271 \\

        HOT-MOE\cite{HOT-MOE}
        & TMM/2025
        & \textcolor{blue}{\textbf{0.5434}}
        & \textcolor{red}{\textbf{0.6934}}
        & 0.7030 & 0.8457
        & 0.4983 & 0.6245 \\

        \midrule

        \textbf{MSP-Net}
        & --
        & \textbf{0.4819}
        & \textbf{0.6222}
        & \textcolor{red}{\textbf{0.7954}}
        & \textcolor{red}{\textbf{0.9428}}
        & \textbf{0.5044}
        & \textcolor{blue}{\textbf{0.6461}} \\

        \bottomrule
    \end{tabular}%
    }

    \vspace{2pt}

    \parbox{\columnwidth}{%
        \centering
        \scriptsize
        Note: Red and blue indicate the best and second-best results, respectively.
    }
\end{table}

\subsection{Comparison on HOT2024 Dataset}
To further explore the performance of MSP-Net, we evaluate on the challenging HOT2024 validation set across VIS, RedNIR, and NIR modalities. As shown in Table~\ref{tab:hot2024_comparison}, MSP-Net performs particularly well under NIR, achieving an AUC of 0.7954 and a DP@20 of 0.9428, which demonstrates strong generalization and resistance to tracking drift in complex heterogeneous spectral scenes. However, its performance under VIS and RedNIR remains behind several competing methods, revealing that limited spectral dimensionality may weaken its high-dimensional feature-mining advantage, while complex spectral routing may interfere with fine-grained spatial alignment. These findings expose the current boundary of MSP-Net under unseen and severely degraded conditions while providing clear directions for improving its adaptability to modality-information loss and extreme long-sequence disturbances.

\begin{table*}[t]
    \centering
    \caption{Attribute-based comparison on the IMEC25 dataset.
    The best and second-best results are highlighted in red and blue.}
    \label{tab:attribute_imec25}

    \begingroup
    \fontsize{7}{8}\selectfont
    \setlength{\tabcolsep}{3.0pt}
    \renewcommand{\arraystretch}{1.08}

    \begin{tabular*}{\textwidth}
    {@{\extracolsep{\fill}}lccccccccccc@{}}
        \toprule
        \textbf{Tracker}
        & \textbf{BC}
        & \textbf{DEF}
        & \textbf{FM}
        & \textbf{IV}
        & \textbf{IPR}
        & \textbf{LR}
        & \textbf{MB}
        & \textbf{OCC}
        & \textbf{OPR}
        & \textbf{OV}
        & \textbf{SV} \\
        \midrule

        SEE-Net\cite{SEE-Net}
        & 0.936
        & 0.962
        & 0.934
        & 0.893
        & 0.913
        & 0.878
        & \textcolor{blue}{\textbf{0.964}}
        & 0.859
        & \textcolor{blue}{\textbf{0.980}}
        & 0.904
        & 0.945 \\

        SiamBAG\cite{SiamBAG}
        & 0.810
        & 0.865
        & 0.762
        & 0.709
        & 0.870
        & 0.864
        & 0.780
        & 0.804
        & 0.866
        & 0.794
        & 0.828 \\

        SENSE\cite{SENSE}
        & 0.937
        & 0.905
        & 0.909
        & 0.907
        & 0.949
        & 0.845
        & 0.814
        & 0.847
        & 0.873
        & \textcolor{blue}{\textbf{0.955}}
        & \textcolor{blue}{\textbf{0.985}} \\

        DASSP\cite{DASSP}
        & 0.940
        & 0.940
        & 0.943
        & 0.933
        & \textcolor{red}{\textbf{0.976}}
        & 0.955
        & 0.925
        & 0.914
        & 0.941
        & 0.930
        & 0.966 \\

        SiamTU\cite{SiamTU}
        & 0.914
        & 0.966
        & 0.931
        & 0.895
        & 0.956
        & 0.917
        & 0.943
        & 0.916
        & \textcolor{blue}{\textbf{0.980}}
        & 0.908
        & 0.959 \\

        MMF-Net\cite{MMF-Net}
        & 0.910
        & 0.866
        & 0.910
        & 0.860
        & 0.944
        & 0.925
        & 0.865
        & 0.865
        & 0.977
        & 0.826
        & 0.952 \\

        SSTrack\cite{SSTRACK}
        & \textcolor{red}{\textbf{0.977}}
        & 0.982
        & 0.951
        & 0.921
        & 0.965
        & 0.932
        & 0.935
        & 0.901
        & 0.959
        & 0.935
        & 0.947 \\

        SPIRIT\cite{SPIRIT}
        & 0.901
        & 0.972
        & 0.917
        & 0.863
        & 0.939
        & 0.948
        & 0.913
        & 0.907
        & 0.974
        & 0.853
        & 0.896 \\

        PHTrack\cite{PHTrack}
        & 0.895
        & 0.945
        & 0.935
        & 0.935
        & 0.885
        & \textcolor{red}{\textbf{0.993}}
        & 0.913
        & 0.893
        & 0.878
        & \textcolor{red}{\textbf{0.976}}
        & 0.919 \\

        SPTrack\cite{SPTrack}
        & 0.938
        & 0.869
        & 0.923
        & 0.880
        & 0.937
        & 0.934
        & 0.918
        & 0.897
        & 0.962
        & 0.870
        & 0.876 \\

        ProFit\cite{ProFiT}
        & 0.931
        & 0.966
        & \textcolor{red}{\textbf{0.967}}
        & 0.945
        & 0.965
        & \textcolor{blue}{\textbf{0.980}}
        & \textcolor{blue}{\textbf{0.964}}
        & \textcolor{blue}{\textbf{0.938}}
        & \textcolor{red}{\textbf{1.000}}
        & \textcolor{blue}{\textbf{0.955}}
        & 0.970 \\

        \midrule

        \textbf{MSP-Net}
        & \textcolor{blue}{\textbf{0.945}}
        & \textcolor{red}{\textbf{0.989}}
        & \textcolor{blue}{\textbf{0.963}}
        & \textcolor{red}{\textbf{0.959}}
        & \textcolor{blue}{\textbf{0.972}}
        & \textbf{0.949}
        & \textcolor{red}{\textbf{0.965}}
        & \textcolor{red}{\textbf{0.971}}
        & \textcolor{blue}{\textbf{0.980}}
        & \textbf{0.942}
        & \textcolor{red}{\textbf{0.995}} \\

        \bottomrule
    \end{tabular*}

    \vspace{2pt}

    \noindent
    \parbox{\textwidth}{%
        \centering
        \fontsize{7}{8}\selectfont
        Note: Red and blue denote the best and second-best results, respectively.
    }

    \endgroup
\end{table*}

\subsection{Comparison on IMEC25 Dataset}
To further examine the universality and domain generalization of MSP-Net beyond the HOT benchmark series, we evaluate it on the independently collected IMEC25 dataset, which features distinct scene distributions and imaging hardware. As shown in Table~\ref{tab:imec25_comparison}, MSP-Net achieves the highest DP@20 score of 0.9725, outperforming the state-of-the-art ProFit method at 0.9670, while its AUC of 0.7381 remains slightly below ProFit’s 0.7540. The performance trends are further visualized in Fig.~\ref{fig:12}. Although MSP-Net does not rank first on both metrics, its consistently competitive performance across this heterogeneous dataset demonstrates strong tracking capability and favorable cross-domain generalization.

\begin{table}[t]
    \centering
    \caption{Comparison with state-of-the-art hyperspectral trackers on the IMEC25 benchmark.}
    \label{tab:imec25_comparison}

    \begingroup
    \fontsize{7}{8}\selectfont
    \setlength{\tabcolsep}{3.0pt}
    \renewcommand{\arraystretch}{1.08}

    \begin{tabular*}{\columnwidth}
    {@{\extracolsep{\fill}}lccc@{}}
        \toprule
        \textbf{Algorithm}
        & \textbf{Journal/Year}
        & \textbf{AUC}
        & \textbf{DP@20} \\
        \midrule

        SEE-Net\cite{SEE-Net}
        & TIP/2023
        & 0.6900
        & 0.9391 \\

        SiamBAG\cite{SiamBAG}
        & TGRS/2023
        & 0.5360
        & 0.8620 \\

        SENSE\cite{SENSE}
        & Inf. Fusion/2024
        & 0.7030
        & 0.9371 \\

        DASSP\cite{DASSP}
        & PR/2025
        & 0.7143
        & 0.9517 \\

        SiamTU\cite{SiamTU}
        & Infrared Phys.
        & 0.6896
        & 0.9395 \\

        MMF-Net\cite{MMF-Net}
        & TGRS/2024
        & 0.7030
        & 0.9227 \\

        SSTrack\cite{SSTRACK}
        & Inf. Fusion/2024
        & 0.6414
        & 0.9547 \\

        SPIRIT\cite{SPIRIT}
        & TGRS/2024
        & 0.6032
        & 0.9290 \\

        PHTrack\cite{PHTrack}
        & TGRS/2024
        & 0.7120
        & 0.9410 \\

        SPTrack\cite{SPTrack}
        & RS/2025
        & 0.5930
        & 0.9209 \\

        CSSTrack\cite{CSSTrack}
        & PR/2025
        & \textcolor{blue}{\textbf{0.7420}}
        & 0.9320 \\

        ProFit\cite{ProFiT}
        & ISPRS/2025
        & \textcolor{red}{\textbf{0.7540}}
        & \textcolor{blue}{\textbf{0.9670}} \\

        DRSST-Net\cite{DRSST}
        & ISPRS/2026
        & 0.7340
        & 0.9170 \\

        \midrule

        \textbf{MSP-Net}
        & --
        & \textbf{0.7381}
        & \textcolor{red}{\textbf{0.9725}} \\

        \bottomrule
    \end{tabular*}

    \vspace{2pt}

    \noindent
    \parbox{\columnwidth}{%
        \centering
        \fontsize{7}{8}\selectfont
        Note: Red and blue indicate the best and second-best results, respectively.
    }

    \endgroup
\end{table}


\begin{figure}[htbp]
    \centering
    \begin{minipage}{0.24\textwidth}
        \centering
        \includegraphics[width=\linewidth]{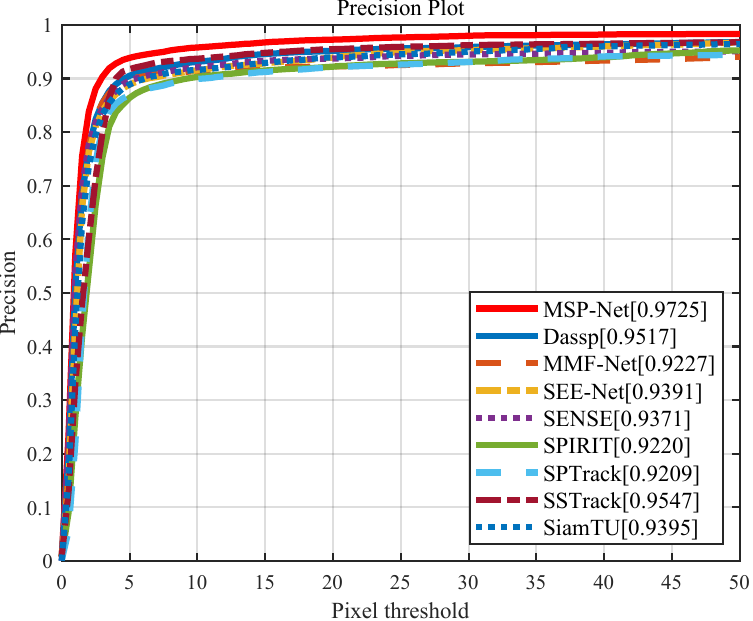}
        \label{fig:12}
    \end{minipage}\hfill
    \begin{minipage}{0.24\textwidth}
        \centering
        \includegraphics[width=\linewidth]{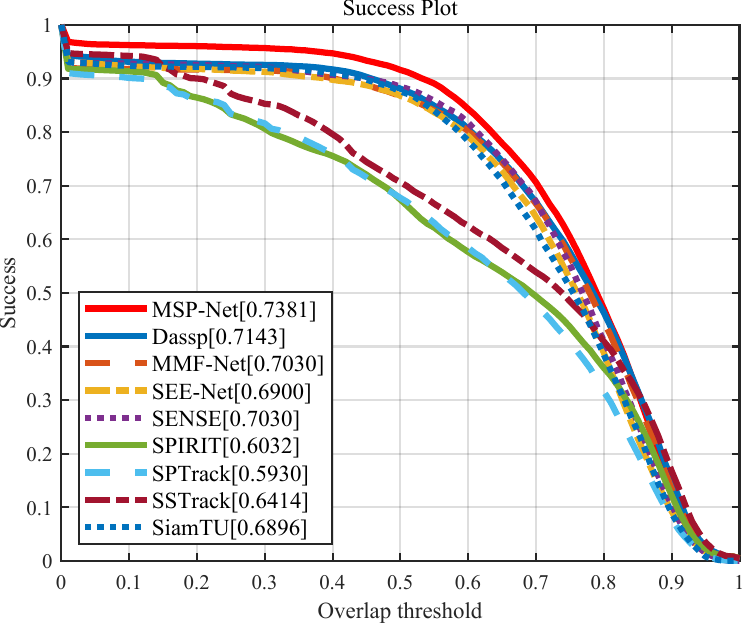}
        \label{fig:12}
    \end{minipage}
    \caption{Precision and success plots of all competing hyperspectral trackers on the IMEC25 dataset.}   
    \label{fig:12}
\end{figure}

According to the attribute-based categorization of the IMEC25 dataset, we further compare the tracking performance under different challenging scenarios using the DP@20 metric. As shown in Table~\ref{tab:attribute_imec25}, MSP-Net ranks among the top three methods across all 11 attributes. Specifically, it achieves the best performance in five scenarios, namely deformation (DEF), illumination variation (IV), motion blur (MB), occlusion (OCC), and scale variation (SV), while ranking second in four scenarios, including background clutter (BC), fast motion (FM), in-plane rotation (IPR), and out-of-plane rotation (OPR). 

\subsection{Ablation Study}



All ablation experiments were conducted at HOT2020 to validate the effectiveness of each component.

\begin{table}[t]
    \centering
    \caption{Ablation study of MSP-Net on the HOT2020 dataset.}
    \label{tab:ablation_mspnet}

    \begingroup
    \fontsize{7}{8}\selectfont
    \setlength{\tabcolsep}{1.0pt}
    \renewcommand{\arraystretch}{1.12}

    \begin{tabular*}{\columnwidth}
    {@{\extracolsep{\fill}}lccccccc@{}}
        \toprule
        \textbf{Model}
        & \textbf{GMSR}
        & \textbf{DSCPM}
        & \textbf{DSCE}
        & \textbf{AUC}
        & $\boldsymbol{\Delta}$\textbf{AUC}
        & \textbf{DP@20}
        & $\boldsymbol{\Delta}$\textbf{DP@20} \\
        \midrule

        Model-1
        & \xmark
        & \xmark
        & \xmark
        & 0.7247
        & --
        & 0.9348
        & -- \\

        Model-2
        & \cmark
        & \xmark
        & \xmark
        & 0.7982
        & \textbf{+7.35\%}
        & 0.9684
        & \textbf{+3.36\%} \\

        Model-3
        & \xmark
        & \cmark
        & \xmark
        & 0.7911
        & \textbf{+6.64\%}
        & 0.9557
        & \textbf{+2.09\%} \\

        Model-4
        & \cmark
        & \cmark
        & \xmark
        & 0.7980
        & \textbf{+7.33\%}
        & 0.9606
        & \textbf{+2.58\%} \\

        Model-5
        & \xmark
        & \cmark
        & \cmark
        & 0.8028
        & \textbf{+7.81\%}
        & 0.9728
        & \textbf{+3.80\%} \\

        \midrule

        \textbf{MSP-Net}
        & \cmark
        & \cmark
        & \cmark
        & \textcolor{red}{\textbf{0.8072}}
        & \textcolor{red}{\textbf{+8.25\%}}
        & \textcolor{red}{\textbf{0.9752}}
        & \textcolor{red}{\textbf{+4.04\%}} \\

        \bottomrule
    \end{tabular*}

    \vspace{2pt}

    \noindent
    \parbox{\columnwidth}{%
        \centering
        \fontsize{7}{8}\selectfont
        Note: Red values indicate the best performance.
    }

    \endgroup
\end{table}

\subsubsection{Effectiveness of the individual proposed method}
Under identical input and backbone settings, we progressively introduced each component into the baseline to evaluate its contribution. As shown in Table~\ref{tab:ablation_mspnet}, GMSR (Model-2) improves AUC and DP@20 by 7.35\% and 3.36\%, respectively, demonstrating the effectiveness of adaptive graph-based spectral routing. DSCPM (Model-3) yields gains of 6.64\% in AUC and 2.09\% in DP@20, confirming that dynamic spatial--spectral prompts provide stronger representations than static prompts. Since DSCE is a temporal updating strategy, it is not evaluated independently. Combining GMSR and DSCPM without DSCE (Model-4) achieves an AUC of 0.7980 and a DP@20 of 0.9606 but shows no clear complementary gain, indicating a mismatch between dynamically evolving spectral features and static prompt conditions. After introducing DSCE (Model-5), AUC further increases by 1.17\%, demonstrating that online condition evolution improves module coordination and reduces representation discrepancies caused by target appearance changes.

\subsubsection{Effectiveness of GMSR}
As shown in Table~\ref{tab:ablation_mspnet}, introducing GMSR significantly improves Model-1. Further ablations in Table~\ref{tab:gmsr_ablation} show that removing manifold graph convolution (\textit{GMSR No Manifold}) reduces AUC and DP@20 by 2.66\% and 3.26\%, confirming the importance of modeling nonlinear inter-band relationships. Replacing hard routing with soft assignment also degrades performance, as soft routing may cause cross-group semantic leakage, whereas hard routing preserves clearer group boundaries. In addition, the node-granularity study in Table~\ref{tab:gmsr_node_ablation} shows that $2 \times 2$ sampling insufficiently represents the manifold topology, while $8 \times 8$ introduces excessive graph coupling. Therefore, the $4 \times 4$ configuration provides the best balance between spectral representation and computational efficiency.

\begin{table}[t]
    \centering
    \caption{Ablation study of the GMSR structure on the HOT2020 dataset.}
    \label{tab:gmsr_ablation}

    \begingroup
    \fontsize{7}{8}\selectfont
    \setlength{\tabcolsep}{3.0pt}
    \renewcommand{\arraystretch}{1.12}

    \begin{tabular*}{\columnwidth}
    {@{\extracolsep{\fill}}lcccc@{}}
        \toprule
        \textbf{Method}
        & \textbf{AUC}
        & $\boldsymbol{\Delta}$\textbf{AUC}
        & \textbf{DP@20}
        & $\boldsymbol{\Delta}$\textbf{DP@20} \\
        \midrule

        MSP-Net
        & \textcolor{red}{\textbf{0.8072}}
        & --
        & \textcolor{red}{\textbf{0.9752}}
        & -- \\

        GMSR No Manifold
        & 0.7806
        & \textbf{-2.66\%}
        & 0.9426
        & \textbf{-3.26\%} \\

        GMSR Soft Routing
        & 0.7958
        & \textbf{-1.14\%}
        & 0.9548
        & \textbf{-2.04\%} \\

        \bottomrule
    \end{tabular*}

    \vspace{2pt}

    \noindent
    \parbox{\columnwidth}{%
        \centering
        \fontsize{7}{8}\selectfont
        Note: Red values indicate the best performance.
    }

    \endgroup
\end{table}



\begin{table}[t]
    \centering
    \caption{Ablation study on the number of manifold nodes in GMSR on the HOT2020 dataset.}
    \label{tab:gmsr_node_ablation}

    \begingroup
    \fontsize{7}{8}\selectfont
    \setlength{\tabcolsep}{3.0pt}
    \renewcommand{\arraystretch}{1.12}

    \begin{tabular*}{\columnwidth}
    {@{\extracolsep{\fill}}lcccc@{}}
        \toprule
        \textbf{Method}
        & \textbf{AUC}
        & $\boldsymbol{\Delta}$\textbf{AUC}
        & \textbf{DP@20}
        & $\boldsymbol{\Delta}$\textbf{DP@20} \\
        \midrule

        \textbf{MSP-Net ($4\times4$)}
        & \textcolor{red}{\textbf{0.8072}}
        & --
        & \textcolor{red}{\textbf{0.9752}}
        & -- \\

        $2\times2$
        & 0.7947
        & \textbf{-1.25\%}
        & 0.9634
        & \textbf{-1.18\%} \\

        $8\times8$
        & 0.7927
        & \textbf{-1.45\%}
        & 0.9636
        & \textbf{-1.16\%} \\

        \bottomrule
    \end{tabular*}

    \vspace{2pt}

    \noindent
    \parbox{\columnwidth}{%
        \centering
        \fontsize{7}{8}\selectfont
        Note: Red values indicate the best performance.
    }

    \endgroup
\end{table}

\begin{figure}[!t]
\centering
\includegraphics[width=3.5in]{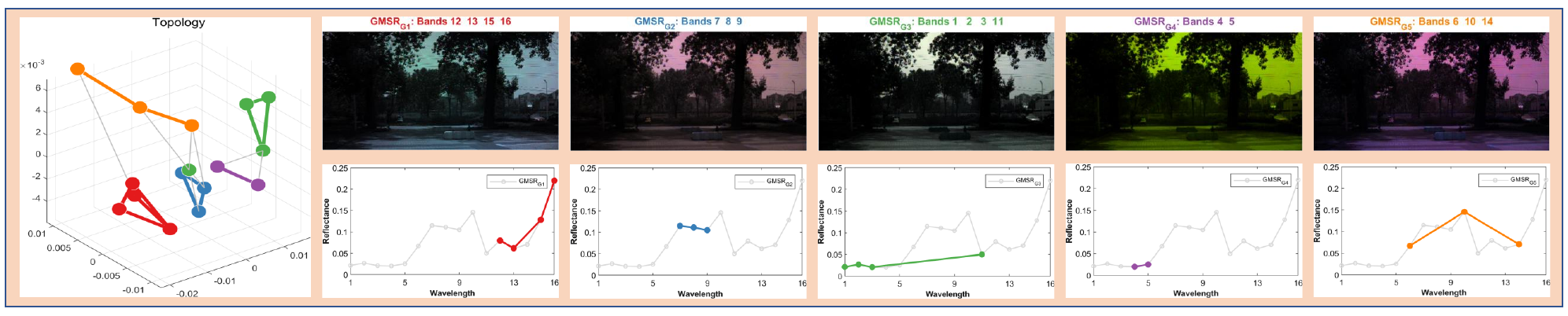}
\caption{Example of cross spectral manifold grouping effect of GMSR model.}
\label{fig:14}
\end{figure}

Fig.\ref{fig:14} presents a visualization of the features processed by GMSR. By constructing a graph convolutional network in the reprojected feature space, GMSR dynamically estimates the manifold relationships among spectral bands. This mechanism relaxes the constraints imposed by physical band indices and adaptively routes non-adjacent but semantically correlated bands across a broad spectral range into the same feature subgroup. Consequently, GMSR suppresses background noise and redundant absorption bands while providing the subsequent ViT backbone with discriminative and decoupled high-order representations.


\subsubsection{Effectiveness of DSCPM}
DSCPM aims to generate scene adaptive prompts from the current spectral context and target appearance while preventing background noise from being indiscriminately amplified. As shown in Table~\ref{tab:dscpm_ablation}, removing dynamic prompts (\textit{No Prompt}) reduces AUC from 0.8072 to 0.7913 and DP@20 from 0.9752 to 0.9654, demonstrating the value of condition-driven prompt learning. More importantly, removing cosine-similarity gating causes larger drops of 2.52\% in AUC and 2.31\% in DP@20. Indicates that effective prompts require not only informative conditions but also selective modulation, allowing DSCPM to enhance target-relevant tokens while suppressing visually similar background regions.

\begin{table}[t]
    \centering
    \caption{Ablation study of the DSCPM structure on the HOT2020 dataset.}
    \label{tab:dscpm_ablation}

    \begingroup
    \fontsize{7}{8}\selectfont
    \setlength{\tabcolsep}{3.0pt}
    \renewcommand{\arraystretch}{1.12}

    \begin{tabular*}{\columnwidth}
    {@{\extracolsep{\fill}}lcccc@{}}
        \toprule
        \textbf{Method}
        & \textbf{AUC}
        & $\boldsymbol{\Delta}$\textbf{AUC}
        & \textbf{DP@20}
        & $\boldsymbol{\Delta}$\textbf{DP@20} \\
        \midrule

        MSP-Net
        & \textcolor{red}{\textbf{0.8072}}
        & --
        & \textcolor{red}{\textbf{0.9752}}
        & -- \\

        DSCPM No Prompt
        & 0.7913
        & \textbf{-1.59\%}
        & 0.9654
        & \textbf{-0.98\%} \\

        DSCPM No CosGate
        & 0.7820
        & \textbf{-2.52\%}
        & 0.9521
        & \textbf{-2.31\%} \\

        \bottomrule
    \end{tabular*}

    \vspace{2pt}

    \noindent
    \parbox{\columnwidth}{%
        \centering
        \fontsize{7}{8}\selectfont
        Note: Red values indicate the best performance.
    }

    \endgroup
\end{table}



\begin{figure}[htbp]
\centering
\includegraphics[width=\columnwidth]{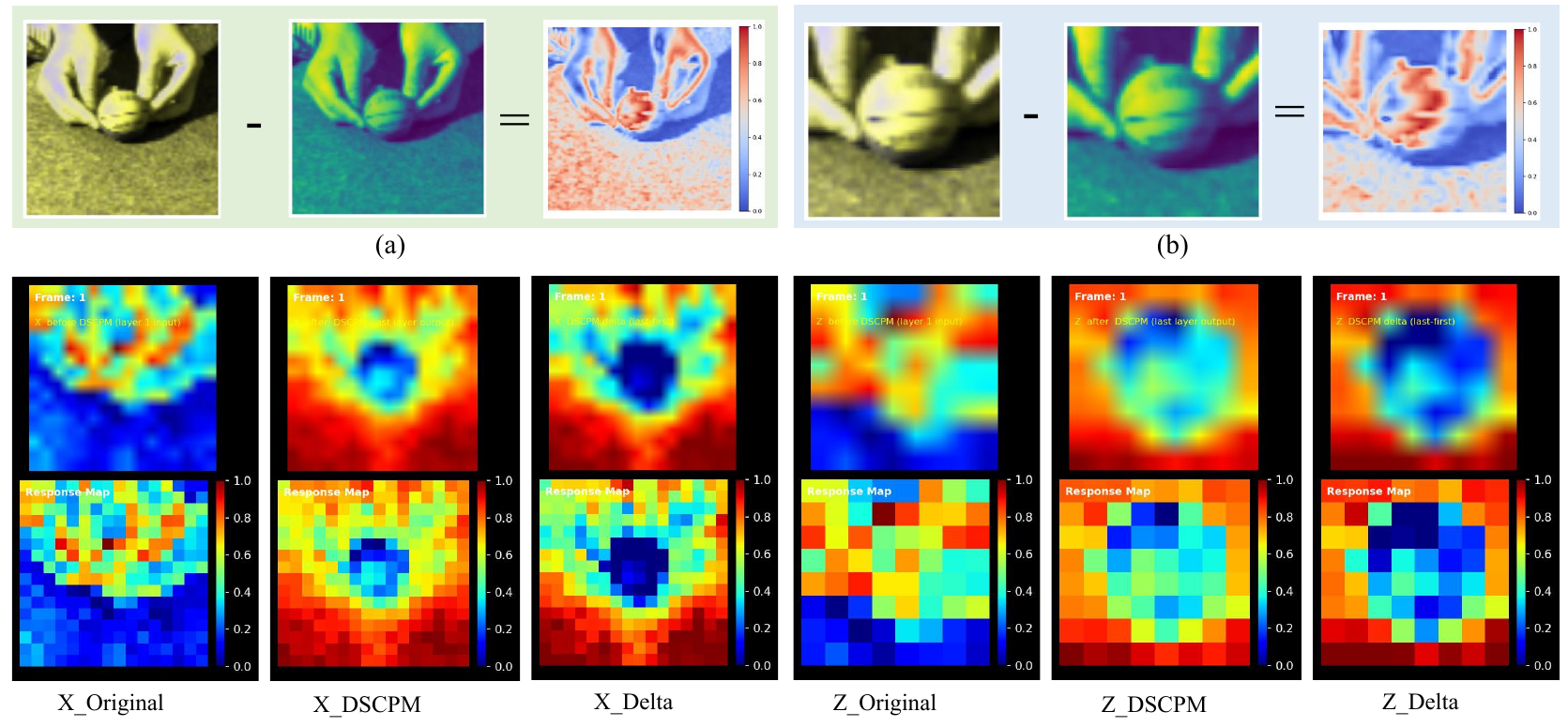}
\caption{Illustration of feature modulation in the DSCPM module. (a) and (b) present the visualizations of the search feature $X$ and the template feature $Z$, respectively. From left to right, the figure compares the feature responses of $X$ and $Z$ before and after DSCPM modulation. $\Delta$ denotes the visualization of the response difference between the DSCPM-modulated features and the original features.}
\label{fig:15}
\end{figure}

Fig.\ref{fig:15} further presents the feature responses of the search and template regions before and after DSCPM modulation. Compared with the original features, the DSCPM responses are more concentrated in target-relevant regions, while the difference maps reveal the non-uniform adjustments applied to the target and background regions. Both the quantitative results and visualizations demonstrate that the performance gains of DSCPM arise not only from dynamic condition-prompt generation but also from the selective control of prompt-injection locations and strengths through cosine-similarity gating.

\subsubsection{Effectiveness of TMMA}
\begin{figure}[htbp]
\centering
\includegraphics[width=\columnwidth]{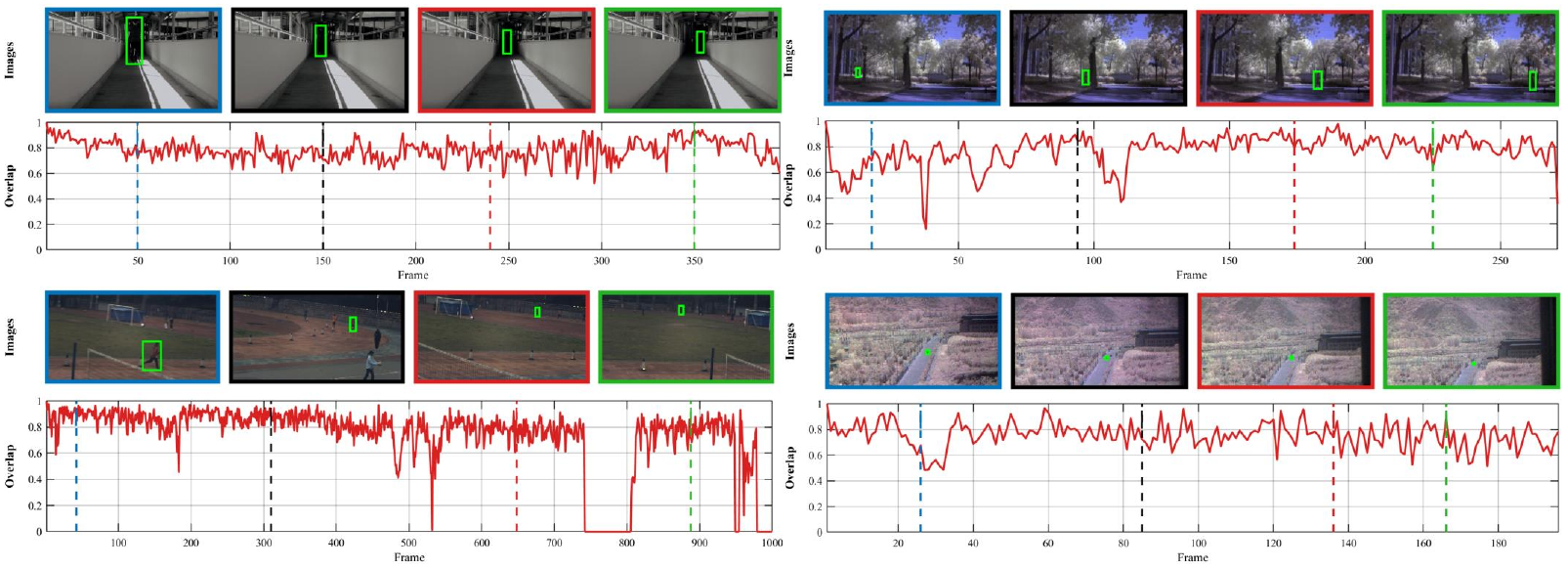}
\caption{MSP Net overlapping tracking results. (a)Student attribute: IV, SV (b) pedestrian7 attribute: BC, OCC, (c) L$\_$runner attribute: OCC, SV (d) motorcycle attribute: DEF, FM, IV, MB, OCC.  The tracking results are marked in green.}
\label{fig:16}
\end{figure}

This experiment evaluates whether TMMA can reduce bounding-box drift and scale jitter caused by distractors, target deformation, and abrupt motion in long-term tracking. As shown in Table~\ref{tab:tmma_ablation}, MSP-Net with TMMA achieves the best AUC and DP@20 of 0.8072 and 0.9752, outperforming the baseline by 8.25\% and 4.04\%, respectively. Removing TMMA decreases its AUC and DP@20 by 0.70\% and 0.71\%, while Model-2, Model-3, and Model-5 show $\Delta\mathrm{AUC}$ values of -1.02\%, -0.43\%, and -0.36\%. Although Model-1 and Model-4 obtain slight gains without TMMA, their absolute performance remains lower. These results suggest that TMMA is most effective when refining high-quality predictions, using reliable historical states to suppress temporal fluctuations without restricting genuine target changes. The resulting overlap trajectories and confidence scores are visualized in Fig.\ref{fig:16}.

\begin{table}[t]
    \centering
    \caption{Ablation study of TMMA on the HOT2020 dataset.}
    \label{tab:tmma_ablation}

    \begingroup
    \fontsize{7}{8}\selectfont
    \setlength{\tabcolsep}{0.8pt}
    \renewcommand{\arraystretch}{1.12}

    \begin{tabular*}{\columnwidth}
    {@{\extracolsep{\fill}}lcccccccc@{}}
        \toprule
        & \multicolumn{4}{c}{\textbf{With TMMA}}
        & \multicolumn{4}{c}{\textbf{No TMMA}} \\
        \cmidrule(lr){2-5}
        \cmidrule(lr){6-9}

        \textbf{Model}
        & \textbf{AUC}
        & $\boldsymbol{\Delta}$\textbf{AUC}
        & \textbf{DP@20}
        & $\boldsymbol{\Delta}$\textbf{DP@20}
        & \textbf{AUC}
        & $\boldsymbol{\Delta}$\textbf{AUC}
        & \textbf{DP@20}
        & $\boldsymbol{\Delta}$\textbf{DP@20} \\
        \midrule

        Model-1
        & 0.7247
        & --
        & 0.9348
        & --
        & 0.7266
        & \textbf{+0.19\%}
        & 0.9383
        & \textbf{+0.35\%} \\

        Model-2
        & 0.7982
        & \textbf{+7.35\%}
        & 0.9684
        & \textbf{+3.36\%}
        & 0.7880
        & \textbf{-1.02\%}
        & 0.9534
        & \textbf{-1.50\%} \\

        Model-3
        & 0.7911
        & \textbf{+6.64\%}
        & 0.9557
        & \textbf{+2.09\%}
        & 0.7868
        & \textbf{-0.43\%}
        & 0.9516
        & \textbf{-0.41\%} \\

        Model-4
        & 0.7980
        & \textbf{+7.33\%}
        & 0.9606
        & \textbf{+2.58\%}
        & \textcolor{red}{\textbf{0.8042}}
        & \textcolor{red}{\textbf{+0.62\%}}
        & \textcolor{red}{\textbf{0.9681}}
        & \textcolor{red}{\textbf{+0.75\%}} \\

        Model-5
        & 0.8028
        & \textbf{+7.81\%}
        & 0.9728
        & \textbf{+3.80\%}
        & 0.7992
        & \textbf{-0.36\%}
        & 0.9652
        & \textbf{-0.76\%} \\

        \midrule

        \textbf{MSP-Net}
        & \textcolor{red}{\textbf{0.8072}}
        & \textcolor{red}{\textbf{+8.25\%}}
        & \textcolor{red}{\textbf{0.9752}}
        & \textcolor{red}{\textbf{+4.04\%}}
        & 0.8002
        & \textbf{-0.70\%}
        & 0.9681
        & \textbf{-0.71\%} \\

        \bottomrule
    \end{tabular*}

    \vspace{2pt}

    \noindent
    \parbox{\columnwidth}{%
        \centering
        \fontsize{7}{8}\selectfont
        Note: Red values indicate the best performance.
    }

    \endgroup
\end{table}

\begin{figure}[htbp]
\centering
\includegraphics[width=3.5in]{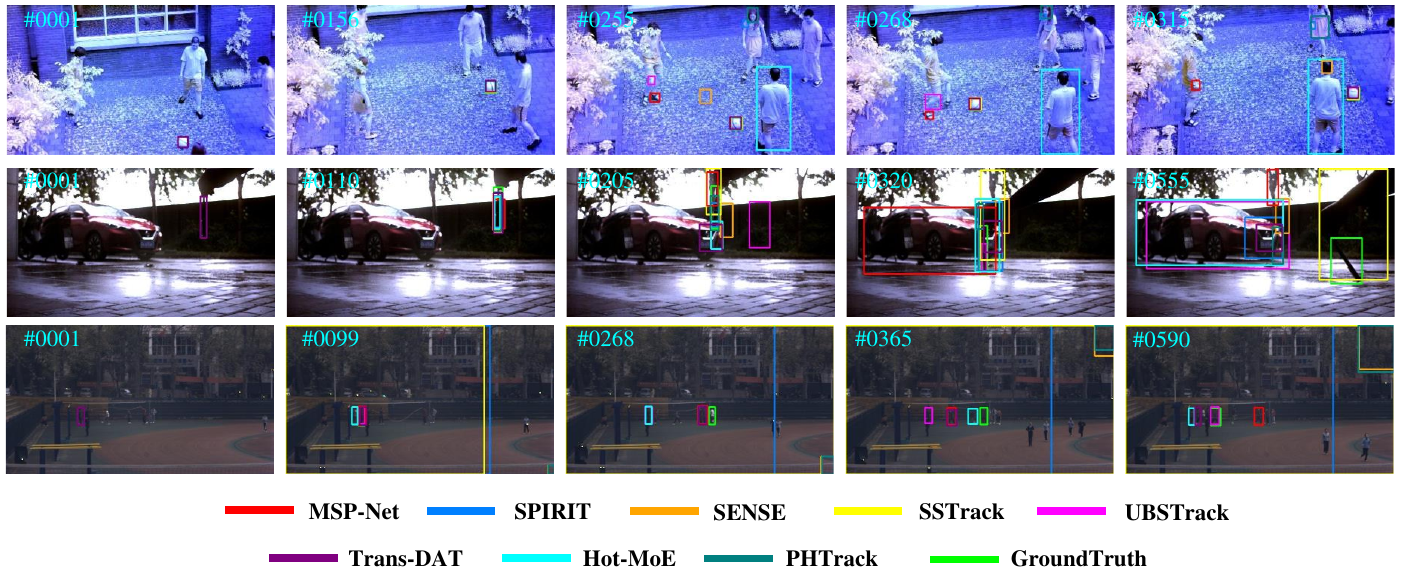}
\caption{Failure case analysis of MSP-Net, scenario is football2, pen, and s$\_$jump2 in the HOT2024 validation set.}
\label{fig:17}
\end{figure}

\subsubsection{Limtations}
To examine the performance limits of MSP-Net, we conduct a failure-case analysis. Despite joint training on HOT2024 and IMEC25, most failures occur in the VIS and RedNIR modalities of the HOT2024 validation set. As shown in Fig.\ref{fig:17}, this is mainly attributed to substantial distribution shifts caused by previously unseen backgrounds, target materials, and sensor-specific spectral response variations. Under such out-of-distribution conditions, the spectral manifold topology may change significantly, making dynamic graph construction less reliable and weakening target-specific spectral representation. As a result, the tracker becomes more vulnerable to deceptive background clutter, abrupt appearance changes, and severe drift. Therefore, improving zero-shot cross-domain generalization to unseen materials and extreme environments remains an important direction for future hyperspectral object tracking research.

\section{Conclusion}
This work aims to overcome tracking drift caused by drastic target appearance variations and interference from similar background clutter in complex hyperspectral scenarios. We propose the MSP-Net, a manifold-aware visual prompt tracking framework that integrates adaptive spectral modeling with temporal feature evolution. Within the framework, GMSR first captures nonlinear relationships among spectral bands through dynamic graph routing, providing informative spectral representations for subsequent processing. Building on these representations, DSCPM combines spectral statistics with template appearance cues to generate target-aware prompts and modulate the backbone features. DSCE further regularizes spectral conditions during training to improve their discriminability and periodically updates them during inference, enabling the prompts to adapt to evolving target appearances without additional parameter optimization. Finally, TMMA exploits temporal morphological cues to stabilize target localization under deformation and abrupt disturbances. Extensive experiments on the HOT series and IMEC25 demonstrate that MSP-Net achieves strong tracking accuracy and robustness across heterogeneous hyperspectral scenarios. Future work will focus on mitigating manifold domain shifts and improving zero-shot generalization to unseen materials and open-world environments.

\bibliographystyle{IEEEtran}
\bibliography{References}   
 
\vspace{80pt}

\vfill

\end{document}